\def\year{2022}\relax
\documentclass[letterpaper]{article} % DO NOT CHANGE THIS
\usepackage{aaai22}  % DO NOT CHANGE THIS
\usepackage{times}  % DO NOT CHANGE THIS
\usepackage{helvet}  % DO NOT CHANGE THIS
\usepackage{courier}  % DO NOT CHANGE THIS
\usepackage[hyphens]{url}  % DO NOT CHANGE THIS
\usepackage{graphicx} % DO NOT CHANGE THIS
\usepackage{natbib}  % DO NOT CHANGE THIS AND DO NOT ADD ANY OPTIONS TO IT
\usepackage{caption} % DO NOT CHANGE THIS AND DO NOT ADD ANY OPTIONS TO IT
\DeclareCaptionStyle{ruled}{labelfont=normalfont,labelsep=colon,strut=off} % DO NOT CHANGE THIS
\usepackage{algorithm}
\usepackage{algorithmic}

\usepackage{newfloat}
\usepackage{listings}
\floatstyle{ruled}
\newfloat{listing}{tb}{lst}{}
\floatname{listing}{Listing}

\newcommand{\beginsupplement}{%
        \setcounter{table}{0}
        \renewcommand{\thetable}{S\arabic{table}}%
        \setcounter{figure}{0}
        \renewcommand{\thesection}{\Alph{section}}
     }

\usepackage{multirow}
\usepackage{amsmath}
\usepackage{amssymb}
\usepackage{amsthm}
\usepackage{booktabs}
\usepackage{ dsfont }
\usepackage{ cite }
\usepackage{float}
\usepackage{soul}
\usepackage{caption}
\usepackage{subcaption}
\usepackage{graphicx}
\usepackage[title]{appendix}
\usepackage{thmtools, thm-restate}
\usepackage{xpatch}
\makeatletter
\xpatchcmd{\@thm}{\thm@headpunct{.}}{\thm@headpunct{}}{}{}
\makeatother

\title{Deep One-Class Classification via Interpolated Gaussian Descriptor}

\author{
    Yuanhong Chen\textsuperscript{\rm 1}\equalcontrib,
    Yu Tian\textsuperscript{\rm 1}\equalcontrib\thanks{Corresponding author.},
    Guansong Pang\textsuperscript{\rm 2}, and
    Gustavo Carneiro\textsuperscript{\rm 1}
}
\affiliations{
\textsuperscript{\rm 1}Australian Institute for Machine Learning, University of Adelaide, Australia\\
\textsuperscript{\rm 2}School of Computing and Information Systems, Singapore Management University, Singapore
}

\begin{document}
\maketitle

\begin{abstract}
% \guansong{`Deep one-class anomaly detection' (in the title) can be misleading, i.e., detecting only one-class anomalies. Change to Deep one-class classification instead?}
% \gustavo{I prefer "Deep Anomaly Detection via Interpolated Gaussian Descriptor  One-class Classifier"}
% \yuanhong{If we want to build a strong connection with DSVDD we can use 'Deep one-class classification via Interpolated Gaussian Descriptor'}\gustavo{I'm ok with that.}
One-class classification (OCC) aims to learn an effective data description to enclose all normal training samples and detect anomalies based on the deviation from the data description. 
Current state-of-the-art OCC models learn a compact normality description by hyper-sphere minimisation, but they often suffer from overfitting the training data, especially when the training set is small or contaminated with anomalous samples. 
%To address this issue, we introduce the interpolated Gaussian descriptor (IGD) method, a novel OCC model that learns a prior-driven one-class Gaussian classifier trained with adversarially interpolated training samples.  
To address this issue, we introduce the interpolated Gaussian descriptor (IGD) method, a novel OCC model that learns a one-class Gaussian anomaly classifier trained with adversarially interpolated training samples.  
%a Gaussian prior-driven one-class hyper-sphere with adversarial interpolation. 
%The imposed prior 
The Gaussian anomaly classifier differentiates the training samples based on their distance to the Gaussian centre and the standard deviation of these distances, offering the model a discriminability w.r.t. the given samples during training. 
The adversarial interpolation is enforced to consistently learn a smooth Gaussian descriptor, even when the training data is small or contaminated with anomalous samples. 
This enables our model to learn the data description based on the representative normal samples rather than fringe or anomalous samples, resulting in significantly improved normality description. 
% The interpolated Gaussian descriptor is learned by an expectation-maximisation (EM) algorithm, where the E-step estimates the Gaussian distribution parameters while the M-step optimises the interpolated Gaussian descriptor.
In extensive experiments on diverse popular benchmarks, including MNIST, Fashion MNIST, CIFAR10, MVTec AD and two medical datasets, IGD achieves better detection accuracy than current state-of-the-art models. IGD also shows better robustness
% than DSVDD and other OCC methods
% DSVDD hasn't been mentioned before
in problems with small or contaminated training sets. 

\end{abstract}

% {-.3cm}
\section{Introduction}
Anomaly detection and segmentation are critical tasks in many real-world applications, such as the identification of defects on industry objects~\cite{mvtecad} or abnormalities from medical images~\cite{anogan,f-anogan}.
Given that most of the training sets available for this task contain only normal images, existing methods are typically formulated as one-class classifiers (OCC)~\cite{venkataramanan2019attention,dsvdd}.
%GUANSONG: the above references may be replaced with survey papers instead if more space is needed
% -- these methods are also commonly referred to as unsupervised anomaly detectors (UAD). %GUANSONG: We don't need to mention UAD at all in this paper; focus on OCCs instead. UAD itself means different things to different researchers in this area.
OCCs aim to first learn a data description of normal samples in the training set and then use a criterion (e.g., distance to the one-class centre~\cite{dsvdd}) to detect and localise anomalies in test samples. %GUANSONG: MAKE A MINOR CHANGE HERE TO GUARANTEE CONSISTENT DESCRIPTION OF OCC MODELS
% Hence, the functionality of OCCs depends on the normal image distribution generalisation and on the anomaly identification criteria. %This sentence seems to be irrelevant w.r.t. our current storyline

\begin{figure}[t!]
  \centering
    \includegraphics[width=0.48\linewidth]{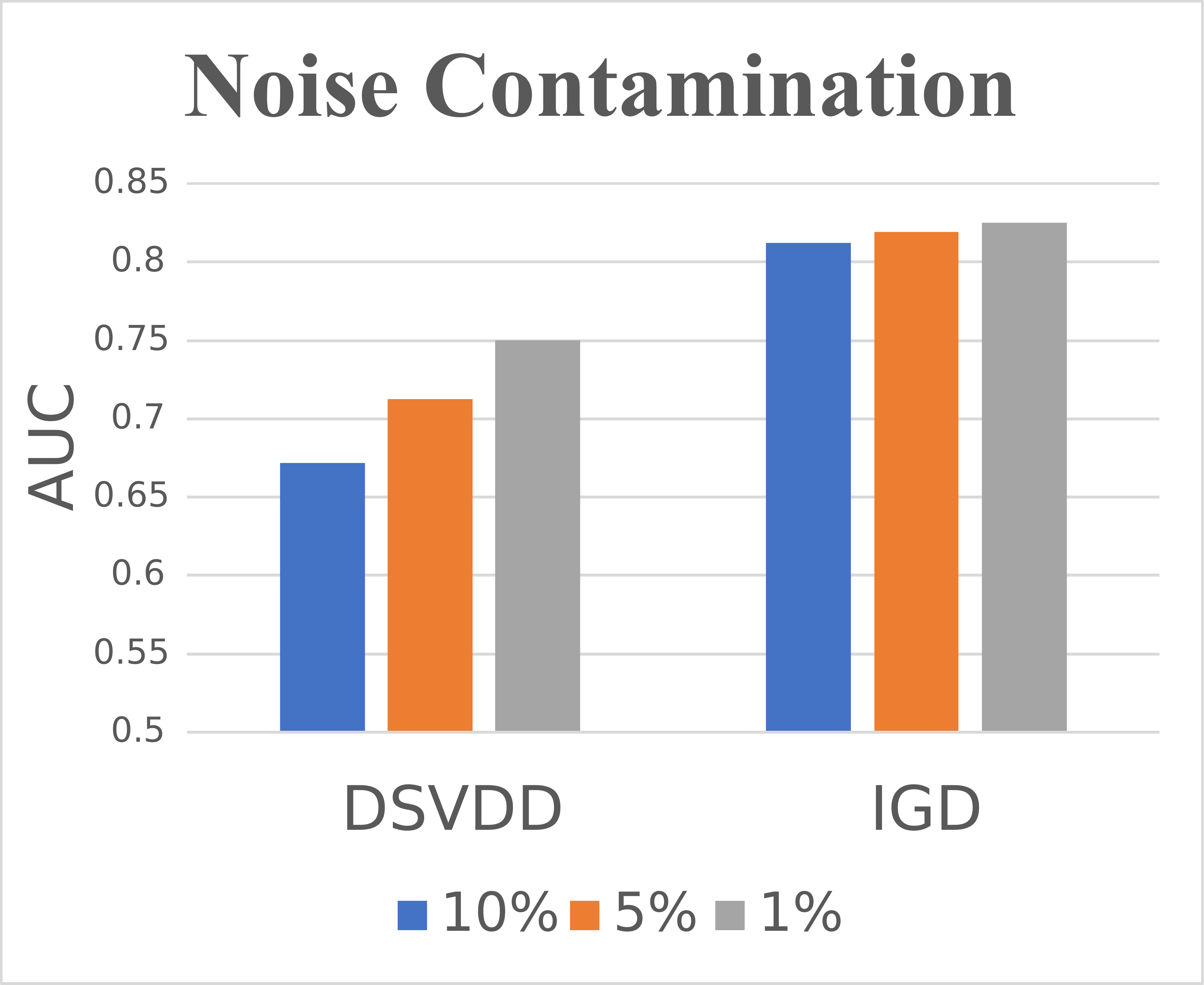}
    \includegraphics[width=0.48\linewidth]{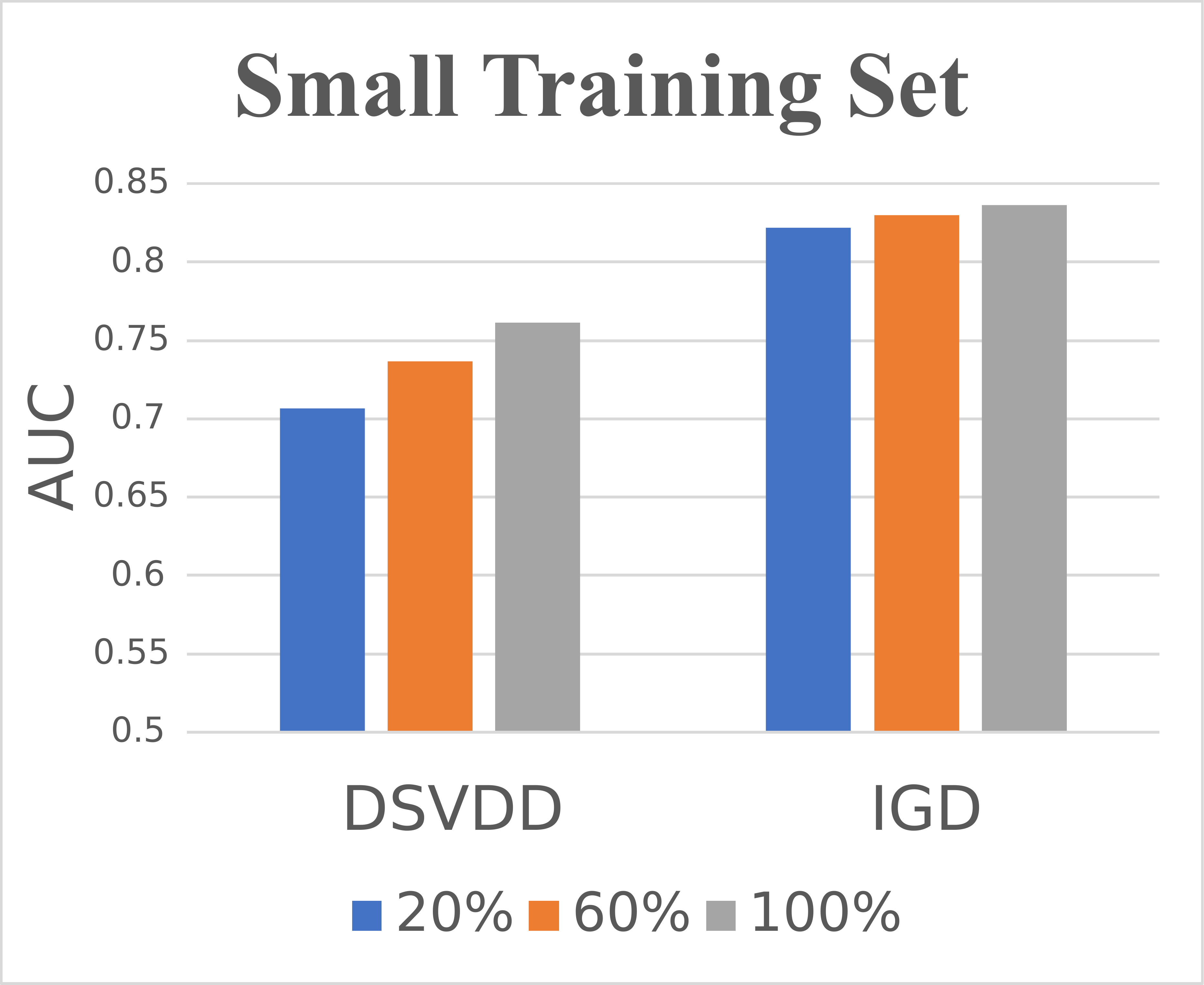}

\caption{Mean testing AUC of DSVDD~\cite{dsvdd}, and our proposed IGD trained with the CIFAR10 training set contaminated with 1\%, 5\% and 10\% of anomalous samples (left), and small training sets, consisting of 20\%, 60\%, and 100\% of the CIFAR10 training set (right).
% \guansong{I suggest to remove DSVDD+REC in this figure, as it is not mentioned in the introduction, or say, it is irrelevant to the storyline we have in the introduction.} 
% \yu{We didn't mentioned much about the reconstruction or the combination. So we may delete the 'DSVDD+REC'. }\guansong{yes, just delete it from this figure. we can reveal more details in the experiments section.}
% \gustavo{Agree -- we should remove it}
% \textbf{Bottom}: t-sne results comparison between the three methods on 'bottle' class of MVTec AD. 
% \guansong{the bottom figure is somehow misleading because anomalies are presumed to be the minority, not the majority as in the figure here.} \gustavo{That's a good point.  We probably need to explain that the the density of the points in the normal cluster is a lot higher. However, we have a problem to explain the DSVDD case, unless normal samples are superimposed in very specific locations.}
% \\
% \gustavo{Why do we have DSVDD+REC on top and GAC at the bottom?} \yu{Yeah, maybe we should be consistent with the top and bottom figure. I prefer to change GAC to DSVDD+REC as the IGD is our main contribution. }
% \yuanhong{The bottom figure aim to show the difference after applying the interpolation originally.} 
% \yuanhong{can we put the bottom one into the appendix? we still need to reduce one and half pages}
}
\label{fig:motivation1}
\end{figure}

State-of-the-art (SOTA) OCC models are trained by minimising 
the radius of a hyper-sphere to enclose all training samples in the representation space~\cite{dsvdd,perera2019learning,ruff2019deep}. 
%GUANSONG: ADD ONE MORE REFERENCE TO STRENGTHEN THIS STATEMENT.
To avoid catastrophic collapse, where all training samples are projected to a single point in the representation space, these OCC models fix the hyper-sphere centre and remove the bias terms from the model. 
%, and overcome the catastrophic collapse by removing the biases terms from the model. This restrict the network to learn richer representation that can express better characteristics of the normal training data, and limited the power of leveraging the pre-trained features from ImageNet or self-supervised learning. 
%Moreover, DSVDD constrain the distribution of normal images to lie at specific regions of the feature space cannot provide a robust generalisation with such scenario
Even though these SOTA OCC models show accurate anomaly detection results in several benchmarks, 
% it has been noticed that 
they can overfit the training data, particularly when the training set is small or contaminated with anomalous samples, as shown by the results of DSVDD~\cite{dsvdd} in Fig.~\ref{fig:motivation1}.

In this paper, we introduce the interpolated Gaussian descriptor (IGD) method to address the overfitting issue presented in SOTA OCC models.
% In particular, IGD learns a prior-driven one-class Gaussian classifier trained with adversarially interpolated training samples, where this prior discriminates training samples based on their distance to the Gaussian centre and the standard deviation of these distances.
IGD is based on a one-class Gaussian anomaly classifier modelled with adversarially interpolated training samples.
The classifier is trained to
build a normality description to
discriminate training samples based on their distance to the Gaussian centre and the standard deviation of these distances.
The smoothness of the normality description is enforced by the adversarial interpolation of the training samples that constrains the training of IGD to be 
%GUANSONG: SHALL WE REMOVE 'TO BE' HERE?
based on representative normal samples rather than fringe or anomalous samples.
This allows the normality description of IGD to be more robust than the SOTA OCC models, particularly when the training set is small or contaminated with anomalous samples, as shown in
Fig.~\ref{fig:motivation1} and t-SNE results in appendix.
% \footnote{The Supp. Material has a t-SNE graph comparing the distribution of normal and anomalous samples of a few methods, which shows the superior normality desccription of IGD.}.  
% \yu{Compared with DSVDD, our proposed IGD based EM alleviates the collapse issue without fixing the centre and removing the bias terms, further boosting the generalisation ability of the model.   } \yu{I want to emphasis that we don't remove bias. Otherwise they may assume we remove the bias as DSVDD..... As we discussed, our main contribution is that we resolve overfitting issues and provide denser hyper-sphere without the collapsing issues. Resolving overfitting and collapsing at the same time was often counter-intuitive in previous methods (i.e., DSVDD) } \guansong{We don't need to emphasize this detail here I think}
% To the best of our knowledge, this paper is the first in computer vision to explore such robustness in anomaly detection problems \guansong{not sure about this statement. it should be some related work somewhere.}.
% Furthermore, the learning of IGD is formulated as an expectation-maximisation (EM) optimisation, which is shown to be theoretically correct and to converge to a stationary point under certain conditions. \guansong{: This sentence is void. We should use the space to make more important statements, for example, to summarize our main contributions. Some ad-hoc reviewers may reject our paper simply due to the lack of the contribution statements.}

In summary, our paper makes the following contributions:
\begin{itemize}
     \item One novel OCC model that targets the learning of an effective normality description based on representative normal samples rather than fringe or anomalous samples, resulting in an improved anomaly classifier, compared with the SOTA; 
     \item One new OCC optimisation approach based on a theoretically sound derivation of the expectation-maximisation (EM) algorithm that optimises a Gaussian anomaly classifier constrained by adversarially interpolated training samples and multi-scale structural and non-structural image reconstruction to enforce a smooth normality description; and
     \item One new OCC benchmark to assess the robustness of anomaly detectors to training sets that are small or contaminated with anomalous samples.
\end{itemize}
% \gustavo{How about this?}\guansong{It reads good. I made some minor changes.}
% \gustavo{I added something about the reconstruction -- please check.}
% \yu{some people also considered their extensive experiments and SOTA results as the contribution.. Should we add it? Two contributions seems a bit weak. } \gustavo{Results are not contribution... but we can add the new benchmark on small and contaminated training sets as a contribution -- what do you think?}\yu{I agree it seems that those two benchmark are newly proposed by our paper. It reads good now... }
% Our method shows the best results
Extensive empirical results on six popular anomaly detection benchmarks for semantic anomaly detection, industrial defect detection,
% , and localisation 
and malignant lesion detection show that our model IGD
% is the first method that 
can generalise well across these diverse application domains and perform consistently better than current SOTA detectors. We also show that IGD is more robust than current OCC approaches
when dealing with small and contaminated training sets.
% on CIFAR10~\cite{krizhevsky2014cifar} and MVTec AD~\cite{mvtecad}. %GUANSONG: I think we don't need to mention that much details at this stage
% The criteria to detect and localise image anomalies are usually based on the reconstruction loss between the original image and its reconstructed image obtained from a generative model fitted to represent the normal training data~\cite{liu2018future,ren2015unsupervised,xu2015learning,ionescu2019object,gong2019memorizing,nguyen2019anomaly,sabokrou2017deep,sabokrou2018adversarially,morais2019learning}.
% Such reconstruction loss generally relies on the mean square error (MSE), which can miss multi-scale structural anomalies~\cite{pangguansong2020,SSIM}.
% Recently, a method based on structural similarity index measure (SSIM)~\cite{SSIM} has been proposed~\cite{ae-ssim} to detect single-scale structural anomalies, but
% anomalies can exhibit abnormality in varying scales.
% Therefore, it is important that the criteria to identify anomalies take into account global and local reconstruction errors, with the local errors considering
% multi-scale structural and non-structural anomalies.

% {-.3cm}
\section{Related Work}

\label{sec:unsupervised_anomaly_detection}
% \yu{We can minimise the content of related work. I don't think reviewer will read it in details. Maybe we replace some references with survey and change the contents a bit.}
Unsupervised anomaly detection (UAD) is generally solved with OCCs~\cite{li2021cutpaste,tian2021constrained,tian2020few,tian2021weakly,dsvdd,bergmann2020uninformed,ocgan,Salehi_2021_CVPR,Wang_2021_CVPR,tian2021selfsupervised,bergman2020classification,golan2018deep,defard2021padim,Zavrtanik_2021_ICCV,wang2016s}.
%thatlearns a distribution of normal samples, and anomalous samples are detected based on how much they deviate from this distribution.
%The first OCC models relied on hand-crafted features~\cite{basharat2008learning,wang2014learning,zhang2009learning}, but recent deep learning models that learn both the feature extractor and classifier tend to be more effective~\cite{dsvdd,bergmann2020uninformed,lsa,ocgan,bergman2020classification,golan2018deep}. 
A representative OCC model is DSVDD~\cite{dsvdd}, which forces normal image features to be inside a hyper-sphere with a pre-defined centre and a radius that is minimised to include all training images. Then, test images that fall inside the  hyper-sphere are classified as normal, and the ones outside are anomalous.
Although powerful, the hard boundary of SVDD can cause the model to overfit the training data -- this problem was tackled with a soft-boundary SVDD~\cite{dsvdd}, but it can still overfit given that it lacks enough generalisation constraints. 
OCC methods can also rely on generative models, such as generative adversarial network (GAN) or Auto-encoder (AE).
In~\cite{ocgan}, a GAN is trained to produce normal samples, and its discriminator is used to detect anomalies, but the complex training process of GANs represents a disadvantage of this approach.
An AE~\cite{ionescu2019object,gong2019memorizing,nguyen2019anomaly,sabokrou2017deep,sabokrou2018adversarially,venkataramanan2019attention} is trained to reconstruct normal data, and the anomaly score is defined as the reconstruction error between the input and reconstructed images.  AE approaches depend on the MSE reconstruction loss, which does not work well for structural anomalies.  Alternatively, single-scale SSIM loss~\cite{ae-ssim} tends to work well for structural anomalies of a specific size, but it may work poorly for non-structural anomalies and structural anomalies outside that specific size. A more detailed review of these methods can be found in \cite{pang2021deep}.

An important aspect of current UAD approaches is their dependence on pre-trained models to produce SOTA results. UAD models can be pre-trained on ImageNet~\cite{venkataramanan2019attention,bergmann2020uninformed} or self-supervised tasks~\cite{golan2018deep,bergman2020classification}.
To allow a fair comparison with current UAD methods, we pre-train IGD with self-supervision and ImageNet.

Unsupervised anomaly localisation targets the segmentation of anomalous image pixels or patches, containing, for example, lesions in medical images~\cite{Li_2019_CVPR}, defects in industry images~\cite{mvtecad,bergmann2020uninformed}, or road anomalies in traffic images~\cite{pathak2015anomaly,tian2021pixel}.
% %, and anomalies in traffic images~\cite{pathak2015anomaly}.
The main idea explored is based on extending the image based OCC to a pixel-based OCC, where testing produces a pixel-wise anomaly score map~\cite{baur2018deep,ae-ssim}. 
% % The way such scores are computed are based on the same anomaly detection criteria presented above in Sec.~\ref{sec:unsupervised_anomaly_detection}.
In general, methods that can localise anomalies~\cite{venkataramanan2019attention, bergmann2020uninformed} are tuned to particular anomaly sizes and structure, which can cause then to miss anomalies outside that range of sizes and structure.  
To avoid this issue, we design IGD to detect multi-scale structural and non-structural anomalies to improve the anomaly localisation accuracy.

\begin{figure*}[t!]
\begin{center}
\includegraphics[width=0.9\linewidth]{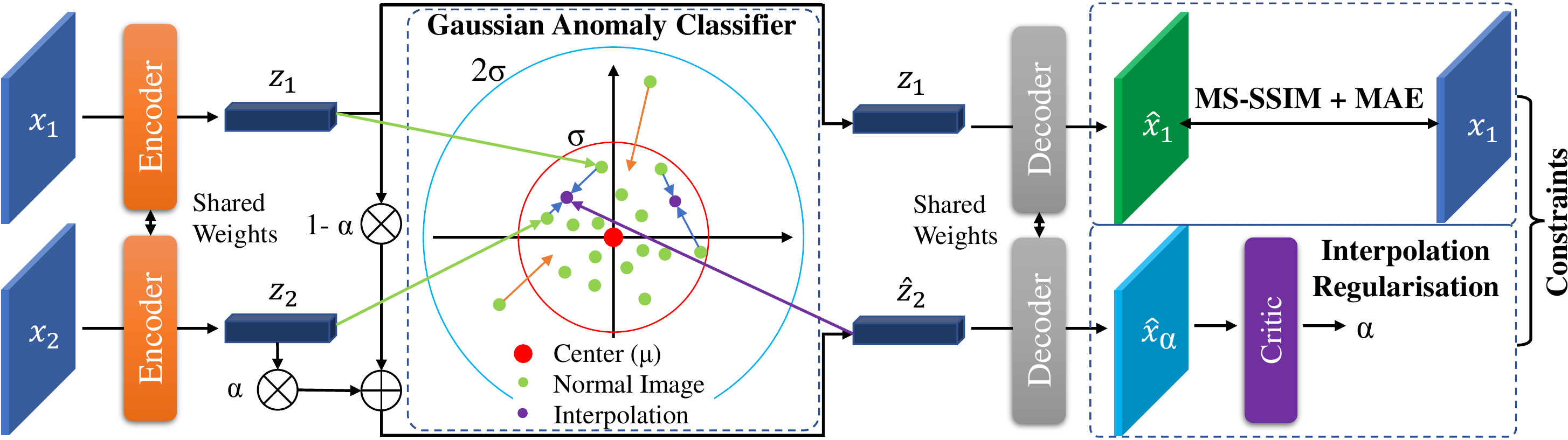}
\end{center}
\caption{
Our IGD consists of an encoder that transforms image $\mathbf{x}$ into representation $\mathbf{z}$, 
a decoder to reconstruct the image (trained with MS-SSIM and MAE losses), 
a Gaussian anomaly classifier trained to push the normal image representation close to the centre of the estimated normal image distribution (denoted by a Gaussian with mean $\mu$ and standard deviation $\sigma$), and a critic module that constrains the likelihood maximisation by predicting the interpolation coefficient $\alpha$ that produces a convex combination of training sample representations. Note that critic is a module similar to a GAN discriminator.
}
\label{fig:train}
\end{figure*}

%-------------------------------------------------------------%
\section{Method}

We denote the training set containing only normal samples by $\mathcal{D} = \{ \mathbf{x}_i \}_{i=1}^{|\mathcal{D}|}$, where $\mathbf{x} \in \mathcal{X} \subset \mathbb{R}^{W \times H \times 3 }$ represents an RGB image of width $W$ and height $H$ and sampled from the distribution of normal images as in $\mathbf{x} \sim \mathcal{P}_{\mathcal{X}}$.  The testing set contains normal and anomalous images, where anomalous images can have segmentation map annotations.  This testing set is defined by
$\mathcal{T} = \{ (\mathbf{x}_i,y_i,\mathbf{b}^{({y_i})}_i \}_{i=1}^{|\mathcal{T}|}$, where $y_i \in \mathcal{Y} = \{0,1 \}$ ($0$ denotes a normal and $1$ denotes an anomalous image), the segmentation map 
with the anomaly is denoted by $\mathbf{b}^{({y_i})}_i \in \{0,1\}^{W \times H}$ (i.e., a pixel-wise anomaly map for image $\mathbf{x}_i$) if $y_i = 1$, and $\mathbf{b}^{({y_i})}_i = 0^{W \times H}$ if $y_i = 0$.

% {-.2cm}
\subsection{Interpolated Gaussian Descriptor (IGD)}
\label{sec:EM}

As depicted in Fig.~\ref{fig:train}, the IGD model is represented by the general classifier $p_{\theta}(y=0|\mathbf{x},\mathcal{P}_{\mathcal{X}})$ that consists of an encoder $\mathbf{z}=f_{\psi}(\mathbf{x})$ that transforms a training sample from the image space $\mathcal{X}$ to a representation space $\mathcal{Z} \in \mathbb{R}^Z$, a Gaussian anomaly classifier $p_{\theta}(y=0|\omega,\mathbf{x}) \in [0,1]$ that takes the normal image distribution parameter $\omega$ and image $\mathbf{x}$ to estimate the probability that it is normal, a decoder $\hat{\mathbf{x}}=g_{\phi}(\mathbf{z})$ that reconstructs an image from the representation space, and a critic module $\alpha=d_{\eta}(g_{\phi}(\alpha \mathbf{z}_1 + (1-\alpha)\mathbf{z}_2))$ that predicts the interpolation constraint parameter $\alpha \in [0,1]$, with $\mathbf{z}_1,\mathbf{z}_2$ obtained from the encoder $f_{\psi}(.)$.
The IGD parameter $\theta \in \Theta$ represents all module parameters $\{ \psi,\phi,\eta \}$ and is estimated with maximum likelihood estimation (MLE):
\begin{equation}
    \theta^* = \arg\max_{\theta}  \frac{1}{|\mathcal{D}|}\sum_{\mathbf{x}_i \in \mathcal{D}} \log p_{\theta}(y_i=0|\mathbf{x}_i,\mathcal{P}_{\mathcal{X}}).
    \label{eq:main_likelihood}
\end{equation}
We train the one-class classifier in~\eqref{eq:main_likelihood} using an
EM optimisation~\citep{dempster1977maximum}, where the mean and standard deviation of the normal image distribution are estimated during the E-step, instead of being explicitly optimised~\cite{dsvdd}, reducing the risk of overfitting.
To encourage the M-step to learn an effective normality description (such that the optimisation is robust to small and contaminated training sets), we add an adversarial interpolation constraint to enforce linear combinations of normal image representations to belong to the normal distribution. 
We further increase the robustness of IGD to overfitting by constraining the optimisation of the M-step to enforce accurate image reconstruction from its representation.
Below, we provide more details about the training process.

To formulate the EM optimisation, we re-write the log-likelihood in~\eqref{eq:main_likelihood} as
\begin{equation}
\begin{split}
    \log p_{\theta}&(y_i=0|\mathbf{x}_i,\mathcal{P}_{\mathcal{X}}) \\ %&=\mathbb{E}_{q(\omega)} \left [ \log \left ( p_{\theta}(y_i=0|\mathbf{x}_i,\mathcal{P}_{\mathcal{X}})\frac{q(\omega)}{q(\omega)} \right ) \right ] \\
    &= \ell_{ELBO}(q,\theta)+KL[q(\omega)||p_{\theta}(\omega|\mathcal{P}_{\mathcal{X}})].
\end{split}
\label{eq:log_likelihood_p_y_x_theta}
\end{equation}
with $\omega \in \mathcal{W} \subset \mathbb{R}^Z \times \mathbb{R}$ denoting the latent variables (mean and standard deviation) that describe the distribution of normal image representations (defined in more detail below).
In~\eqref{eq:log_likelihood_p_y_x_theta}, we remove the conditional dependence of $p_{\theta}(\omega|\mathcal{P}_{\mathcal{X}})$ on $y_i=0$ and $\mathbf{x}_i$ because
$\omega$ is a variable for the whole training distribution defined as
\begin{equation}
    p_{\theta}(\omega|\mathcal{P}_{\mathcal{X}}) = \delta_a(\| \omega(1) - \mu_x\|_2)\delta_a(\omega(2) - \sigma_x),
    \label{eq:p_omega_given_x}
\end{equation}
where $\delta_a(b)=\frac{1}{|a|\sqrt{\pi}}\exp{-(b/a)^2}$ ($a \to 0$ approximates a Dirac delta function, and $a \to \infty$ approximates a uniform function), $\mu_{x} = \mathbb{E}_{\mathbf{x} \sim \mathcal{P}_{\mathcal{X}}}[f_{\psi}(\mathbf{x})]$ and 
$\sigma^2_{x} = \mathbb{E}_{\mathbf{x} \sim \mathcal{P}_{\mathcal{X}}}[\|f_{\psi}(\mathbf{x})-\mu_{x}\|_2^2]$,
with $f_{\psi}(.)$ representing the  encoder; and in~\eqref{eq:log_likelihood_p_y_x_theta}, we also have
%that is a redundant dependence since samples $\mathbf{x}_i \sim \mathcal{P}_{\mathcal{X}}$ are all from $y_i=0$, and 
\begin{equation}
\begin{split}
\ell_{ELBO}&(q,\theta) = \\ &\mathbb{E}_{q(\omega)}[\log p_{\theta}(y_i=0,\omega|\mathbf{x}_i,\mathcal{P}_{\mathcal{X}})]-\mathbb{E}_{q(\omega)}[\log q(\omega) ],   
\end{split}
\label{eq:elbo}
\end{equation}
 where 
 $KL[\cdot]$ denotes the Kullback-Leibler divergence, and $q(\omega)$ represents the variational distribution that approximates $p_{\theta}(\omega|\mathcal{P}_{\mathcal{X}})$, defined in~\eqref{eq:p_omega_given_x}.

The E-step of the EM optimisation zeroes the KL divergence in~\eqref{eq:log_likelihood_p_y_x_theta} by setting 
$q(\omega)=p_{\theta^{old}}(\omega|\mathcal{P}_{\mathcal{X}})$, where $\theta^{old}$ represents the previous EM iteration parameter value. In practice, the E-step sets $\omega(1)$ to $\mu_x$ and $\omega(2)$ to $\sigma_x$, defined in~\eqref{eq:p_omega_given_x}.
Next, the M-step maximises $\ell_{ELBO}$ in~\eqref{eq:elbo}, with:
\begin{equation}
\begin{split}
    \theta^{\star}  =  \arg\max_{\theta}   \frac{1}{|\mathcal{D}|}  \sum_{\mathbf{x}_i \in \mathcal{D}} 
    \Big ( \mathbb{E}_{q(\omega)} [ & \log p_{\theta}(y_i=0|\omega,\mathbf{x}_i) \\
    & + \log p_{\theta}(\omega|\mathcal{P}_{\mathcal{X}}) \big ] \Big ) ,
    \end{split}
    \label{eq:M_step}
\end{equation}
where $\mathbb{E}_{q(\omega)}[\log(q(\omega))]$ is removed from $\ell_{ELBO}$ because it depends only on the previous iteration parameter $\theta^{old}$,
$q(\omega)$ is defined in the E-step above, and 
the conditional dependence of $p_{\theta}(y=0|\omega,\mathbf{x}_i)$ on $\mathcal{P}_{\mathcal{X}}$ is removed because the information from that distribution is summarised in $\theta$.  Therefore, \eqref{eq:M_step} has two components: 1) the classification term represented by the Gaussian anomaly classifier $p_{\theta}(y=0 | \omega, \mathbf{x}_i) = \exp \left( -\frac{\|f_{\psi}(\mathbf{x}) - \omega(1)\|_2^2}{\omega(2)^2} \right)$, with mean $\omega(1)$ and standard deviation $\omega(2)$; and 2) $p_{\theta}(\omega | \mathcal{P}_{\mathcal{X}})$ defined in~\eqref{eq:p_omega_given_x}, 
which approximates a uniform distribution to prevent the confirmation bias of the estimated $\mu_x$ and $\sigma_x$ from~\eqref{eq:p_omega_given_x}.
To promote an effective normality description of IGD, we constrain the M-step~\eqref{eq:M_step} as follows:
\begin{equation}
    \begin{aligned}
\max_{\theta} \quad & \frac{1}{|\mathcal{D}|}\sum_{\mathbf{x}_i \in \mathcal{D}} \mathbb{E}_{q(\omega)}[\log(p_{\theta}(y=0 | \omega, \mathbf{x}_i))]\\
\textrm{s.t.} \quad & \ell_d(\mathbf{x}_i,\theta) = 0, \forall \mathbf{x}_i \in \mathcal{D},\\
  & \ell_{f,g}(\mathbf{x}_i,\theta) = 0, \forall \mathbf{x}_i \in \mathcal{D},
\end{aligned}
\label{eq:optimisation_IGD}
\end{equation}
where $\ell_d(.)$ is a constraint, defined in~\eqref{eq:loss_critic}, to enforce the adversarial linear interpolation of normal image representations to belong to the normal representation distribution, and $\ell_{f,g}(.)$ is a constraint, defined in~\eqref{eq:loss_AE}, to enforce accurate structural and non-structural multi-scale image reconstruction.
Note that the maximisation in~\eqref{eq:optimisation_IGD} constrains the optimisation in~\eqref{eq:M_step}, which means that we are maximising a lower bound to the original M-step. Using Lagrange multipliers, the optimisation in~\eqref{eq:optimisation_IGD} is reformulated to minimise the following loss function:
\begin{equation}
\small
    \ell(\theta,\omega,\mathcal{D}) = \frac{1}{|\mathcal{D}|}
    \sum_{i=1}^{|\mathcal{D}|} 
    \ell_h(\mathbf{x}_i,\omega,\theta) + \lambda_1\ell_d(\mathbf{x}_i,\theta) + \lambda_2\ell_{f,g}(\mathbf{x}_i,\theta),
    \label{eq:loss}
\end{equation}
where 
\begin{equation}
    \ell_h(\mathbf{x},\omega,\theta) = 1-p_{\theta}(y=0 | \omega, \mathbf{x})=p_{\theta}(y=1 | \omega, \mathbf{x}),
    \label{eq:loss_GSVDD}
\end{equation}  
with $p_{\theta}(y=0 | \omega, \mathbf{x})$ defined in~\eqref{eq:M_step}, and $\lambda_1,\lambda_2$ denoting the Lagrange multipliers.
The interpolation constrain $\ell_d(.)$ in~\eqref{eq:optimisation_IGD} and~\eqref{eq:loss} regularises the  training by linearly interpolating the  representations from training images, and estimating the interpolation coefficient with the critic network~\cite{berthelot2018understanding}.
This interpolation constrains
 the normal image distribution denser in the representation space, reducing the likelihood that anomalous representations may land in the same region of the representation space occupied by normal samples.
Unlike Mix-up~\cite{zhang2017mixup}, our interpolation constraint is a self-supervised method that does not rely on data augmentation on the input space and does not interpolate training labels, making it more adequate for our problem because it enforces a compact and dense distribution of normal samples to be estimated for the Gaussian anomaly classifier. 
The critic network is represented by
\begin{equation}
\hat{\alpha} = d_{\eta}\left(\hat{\mathbf{x}}_{\alpha}\right),
\label{eq:critic}
\end{equation}
where $\hat{\mathbf{x}}_{\alpha}=g_{\phi}\left(\alpha \mathbf{z}_1 + (1-\alpha)\mathbf{z}_2\right)$ 
represents the reconstruction of the interpolation of $\mathbf{z}_1=f_{\psi}(\mathbf{x}_1)$ and $\mathbf{z}_2=f_{\psi}(\mathbf{x}_2)$ (with $\alpha \sim \mathcal{U}(0,0.5)$, $\mathbf{x}_1,\mathbf{x}_2 \in \mathcal{D}$,  $\mathbf{x}_1 \ne \mathbf{x}_2$, and $\mathcal{U}$ denoting a uniform distribution )~\cite{berthelot2018understanding}, and $g_{\phi}(.)$ denotes the decoder. The goal of the critic network $d_{\eta}(.)$ is to predict the interpolation coefficient $\alpha$. 
The critic network in~\eqref{eq:critic} is similar to the discriminator in GAN~\cite{gan}, and relies on the following adversarial loss to be optimised~\cite{berthelot2018understanding}
\begin{equation}
    \ell_d(\mathbf{x},\theta) = \|d_{\eta}(\hat{\mathbf{x}}_{\alpha}) - \alpha \|_2^2 + \| d_{\eta}(\hat{\mathbf{x}}_{\zeta})\|_2^2,
    \label{eq:loss_critic}
\end{equation}
where 
$\hat{\mathbf{x}}_{\alpha}$ is defined in~\eqref{eq:critic}, and
$\hat{\mathbf{x}}_{\zeta}=\zeta\mathbf{x}+(1-\zeta)\hat{\mathbf{x}}$, with $\zeta \sim \mathcal{U}(0,1)$
and $\hat{\mathbf{x}}$ denoting a reconstruction of $\mathbf{x}$ by the auto-encoder.
The first term of \eqref{eq:loss_critic} minimises the critic's prediction error for $\alpha$ and the second term regularises the training to ensure that the critic predicts $\hat{\alpha}=0$ when the original image is interpolated with its own reconstruction in the image space $\mathcal{X}$. 

The image reconstruction constrain $\ell_{f,g}(.)$ in~\eqref{eq:optimisation_IGD} and~\eqref{eq:loss} is defined as
\begin{equation}
    \ell_{f,g}(\mathbf{x},\theta) = \ell_{r}(\mathbf{x},\hat{\mathbf{x}},\theta) + \lambda_3 \| d_{\eta}(\hat{\mathbf{x}}_{\alpha}) \|_2^2,
    \label{eq:loss_AE}
\end{equation}
where $\hat{\mathbf{x}}$ is a reconstruction of $\mathbf{x}$ by the auto-encoder, with the image reconstruction loss $\ell_{r}(.)$ to be defined below  in~\eqref{eq:global_reconstruction_loss}, and
$\lambda_3$ is a hyperparameter to weight the regularisation term. This regularisation fools the critic to output $\hat{\alpha}=0$ for interpolated embeddings, independently of $\alpha$, following standard adversarial training~\cite{gan}.
In~\eqref{eq:loss_AE}, we also have
\begin{equation}
\begin{split}
    % \ell_{r}&(\mathbf{x},\hat{\mathbf{x}},\theta) 
    \ell_{r}&(\mathbf{x},\hat{\mathbf{x}},\theta) = \\&\sum_{\omega\in\Omega}
    \rho | \mathbf{x}({\omega}) - \hat{\mathbf{x}}({\omega}) | + (1-\rho)\Big(1-m\big (\mathbf{x}({\omega}),\hat{\mathbf{x}}({\omega})\big )\Big),
    \label{eq:global_reconstruction_loss}
\end{split}
\end{equation}
with $\Omega$ denoting the image lattice, $\rho\in[0,1]$,
$| \mathbf{x}({\omega}) - \hat{\mathbf{x}}({\omega}) |$ representing the MAE loss, and
$m(\mathbf{x}({\omega}),\hat{\mathbf{x}}({\omega})) \in [0,1]$ being the MS-SSIM score~\cite{MS-SSIM}, with larger values indicating higher similarity between patches $\omega \in \Omega$ of the original and reconstructed images. 
Please see details on how to compute the MS-SSIM score in the Supp. Material.

The loss in~\eqref{eq:loss} is used to train two models (see 'Global and Local IGD Models' section in the Supp. Material). A global model that works on the whole image $\mathbf{x}$, and a local model that works on image patches $\mathbf{x}^{(L)}({\omega}) \in \mathcal{X}^{(L)} \subset \mathbb{R}^{W^{(L)} \times H^{(L)} \times 3}$, with $W^{(L)} < W$ and $H^{(L)} < H$, centred at pixel $\omega\in\Omega$ ($\Omega$ is the image lattice).
During inference, the results from the global and local models are combined to produce multi-scale anomaly detection and localisation. Please see the Supp. Material for a visual example of the results produced by the global and local models.
\subsection{Theoretical Guarantees}

 IGD maximises a constrained $\ell_{ELBO}(q,\theta)$ in \eqref{eq:optimisation_IGD} rather than maximising $p_{\theta}(y=0|\mathbf{x},\mathcal{P}_{\mathcal{X}})$ in \eqref{eq:main_likelihood}.  
 Using Theorem 1 in~\citep{dempster1977maximum}, Lemma~\ref{thm:correctness} demonstrates the correctness of IGD, where an increase to the constrained $\ell_{ELBO}(q,\theta)$ implies an increase to $p_{\theta}(y=0|\mathbf{x},\mathcal{P}_{\mathcal{X}})$.
Using Theorem 2 in~\citep{dempster1977maximum}, Lemma~\ref{thm:convergence} proves the convergence conditions of IGD. 

\begin{restatable}[]{lemma}{Correctness}
    \label{thm:correctness}
    Assuming that the maximisation of the constrained $\ell_{ELBO}$ in~\eqref{eq:optimisation_IGD} produces $\theta$ that makes\\ 
    \scalebox{0.87}{
    $\mathbb{E}_{q(\omega)}[\log p_{\theta}(y=0,\omega|\mathbf{x},\mathcal{P}_{\mathcal{X}})] \ge 
    \mathbb{E}_{q(\omega)}[\log p_{\theta^{old}}(y=0,\omega|\mathbf{x},\mathcal{P}_{\mathcal{X}})]$,} \\we have that 
    \scalebox{0.87}{$\left(\log p_{\theta}(y=0|\mathbf{x},\mathcal{P}_{\mathcal{X}}) - \log p_{\theta^{old}}(y=0|\mathbf{x},\mathcal{P}_{\mathcal{X}})\right)$} 
    is lower bounded by\\
    \scalebox{0.8}{
    $\left(\mathbb{E}_{q(\omega)}[\log p_{\theta}(y=0,\omega|\mathbf{x},\mathcal{P}_{\mathcal{X}})] - 
    \mathbb{E}_{q(\omega)}[\log p_{\theta^{old}}(y=0,\omega|\mathbf{x},\mathcal{P}_{\mathcal{X}})]\right) \ge 0$,} \\
    with $q(\omega)=p_{\theta^{old}}(\omega|\mathcal{P}_{\mathcal{X}})$.
 \end{restatable}
 
\begin{proof}
% Please see proof in~\appendixname~\ref{sec:correctness_analysis_proof}.
Please see proof in Supp. Material.
\end{proof}

\begin{restatable}[]{lemma}{Convergence}
    \label{thm:convergence}
    Assume that $\{ \theta^{(e)}\}_{e=1}^{+\infty}$ denotes the sequence of trained model parameters from the constrained optimisation of $\ell_{ELBO}$ in~\eqref{eq:optimisation_IGD} such that: 1) the sequence $\{ \log p_{\theta^{(e)}}(y=0|\mathbf{x},\mathcal{P}_{\mathcal{X}})\}_{e=1}^{+\infty}$ is bounded above, and 2) 
    \scalebox{0.79}{$\left(\mathbb{E}_{q(\omega)}[\log p_{\theta^{(e+1)}}(y=0,\omega|\mathbf{x},\mathcal{P}_{\mathcal{X}})] - 
    \mathbb{E}_{q(\omega)}[\log p_{\theta^{(e)}}(y=0,\omega|\mathbf{x},\mathcal{P}_{\mathcal{X}})]\right) \ge $} \\
    \scalebox{0.8}{
    $\xi \left(\theta^{(e+1)}-\theta^{(e)}\right)^{\top}\left(\theta^{(e+1)}-\theta^{(e)}\right)$}, 
    for $\xi>0$ and all $e \ge 1$, and $q(\omega)=p_{\theta^{(e)}}(\omega|\mathcal{P}_{\mathcal{X}})$.  
    Then $\{\theta^{(e)}\}_{e=1}^{+\infty}$ converges to some $\theta^{\star} \in \Theta$.
 \end{restatable}
 
\begin{proof}
% Please see proof in~\appendixname~\ref{sec:convergence_analysis_proof}.
Please see proof in Supp. Material.
\end{proof}

\begin{figure}[htp]
    \centering
    \includegraphics[width=\linewidth]{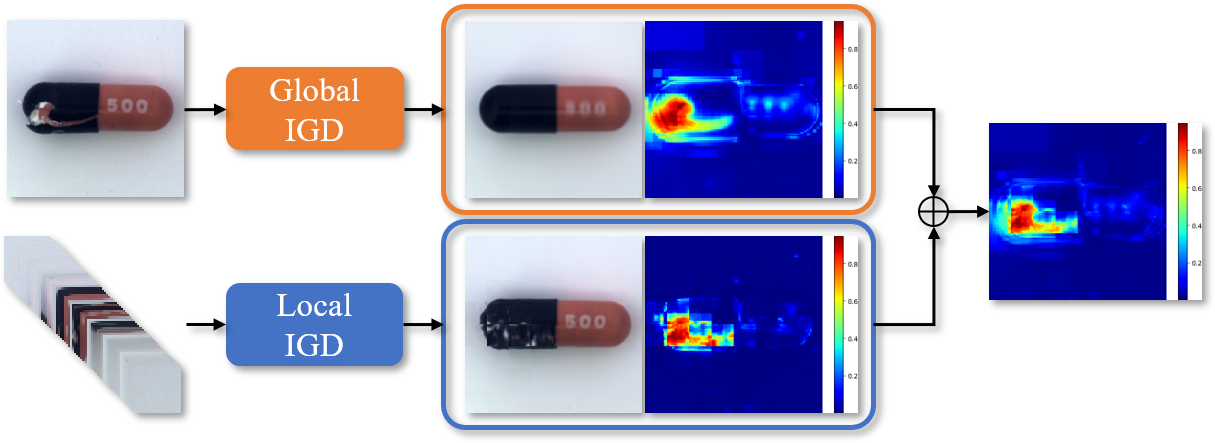}
   \caption{Example of the multi-scale structural and non-structural anomaly localisation result for an MVTec AD~\cite{mvtecad} image, using both the local and global IGD models. The global model tends to produce smooth results but with some mistakes, while the local model produces jagged results, but without the global mistakes, so by combining the two results, we obtain a smooth and correct anomaly heatmap.  %\yu{change figure to Global IGD and Local IGD}
   %\yuanhong{Can we move this figure to appendix?} \yu{If we don't have space,  I think we can. }
   }
    \label{fig:multi-test}
\end{figure}

% {-.2cm}
\subsection{Training and Inference}

The global and local IGD models are trained separately (see Fig.~\ref{fig:multi-test}), following the EM optimisation, where the E-step estimates the the latent variable $\omega$ in~\eqref{eq:p_omega_given_x}, and the M-step minimises the loss in~\eqref{eq:loss} to obtain $\theta^{\star}$.

During inference, \textbf{anomaly detection} is performed by combining the global and local IGD anomaly scores for a testing image $\mathbf{x}$ as in:
\begin{equation}
    s(\mathbf{x}) = s^{(G)}(\mathbf{x}) + s^{(L)}(\mathbf{x}).
    \label{eq:anomaly_score_image}
\end{equation}
% \yuanhong{We can remove $\beta$ accurately, because it is set to 0.5 all the time.}
The global score in~\eqref{eq:anomaly_score_image} is defined as
\begin{equation}
    s^{(G)}(\mathbf{x}) = \ell^{(G)}_{r}(\mathbf{x},\hat{\mathbf{x}},\theta^*) + \ell^{(G)}_h(\mathbf{x},\theta^*),
    \label{eq:global_score_detection}
\end{equation}
% where $\xi \in [0,1]$ weights the contribution between 
where $\ell^{(G)}_{r}(.)$ denotes the reconstruction loss from~\eqref{eq:global_reconstruction_loss} and $\ell^{(G)}_h(.)$ denotes the Gaussian anomaly classification loss from~\eqref{eq:loss_GSVDD} (both computed with the global IGD model using the whole images), and $\hat{\mathbf{x}}$ is the reconstruction of $\mathbf{x}$ produced by the auto-encoder.  The local score in~\eqref{eq:anomaly_score_image} is defined as
\begin{equation}
\begin{split}
s^{(L)}(\mathbf{x})=\max_{\omega\in\Omega} \Big( &
 \ell^{(L)}_{r}\big(\mathbf{x}^{(L)}({\omega}),\hat{\mathbf{x}}^{(L)}({\omega}),\theta^*\big)+ \\
& \ell^{(L)}_h\big(\mathbf{x}^{(L)}({\omega}),\theta^* \big) \Big),
\end{split}
\label{eq:local_score_detection}
\end{equation}
% \yuanhong{$\mathbf{x}^{(L)}\langle{\omega}\rangle>$}
where $\ell^{(L)}_{r}(.)$ and $\ell^{(L)}_h(.)$ are the reconstruction and Gaussian anomaly classification losses computed from the local model, with $\mathbf{x}^{(L)}({\omega})$ denoting an image patch of size $W^{(L)}\times H^{(L)} \times 3$ at pixel $\omega \in \Omega$.
The use of max pooling of the local scores in~\eqref{eq:local_score_detection} facilitates detection of images that contain anomalies covering a small region of the image.
\textbf{Anomaly localisation} is computed for each pixel $\omega \in \Omega$ to produce a local score   
\begin{equation}
\begin{split}
    l(\mathbf{x}({\omega})) = &
      \ell^{(G)}_{r}\big (\mathbf{x}(\omega),\hat{\mathbf{x}}(\omega),\theta^* \big ) + \\
     &\ell^{(L)}_{r}\big (\mathbf{x}^{(L)}({\omega}),\hat{\mathbf{x}}^{(L)}({\omega}),\theta^* \big ),
\end{split}
    \label{eq:localisation_loss^(G)lobal_local_MS-SSIM}
\end{equation}
with
\begin{equation}
\begin{split}
    \ell^{(G)}_{r}\big (\mathbf{x}(\omega), & \hat{\mathbf{x}}(\omega),\theta^* \big ) =  \rho \big| \mathbf{x}(\omega) - \hat{\mathbf{x}}(\omega) \big |  +\\ &(1-\rho) \Big (1-m^{(G)}\big (\mathbf{x}(\omega),\hat{\mathbf{x}}(\omega)\big)\Big ),
    \end{split}
    \label{eq:localisation_map}
\end{equation}
where $\rho$ and $m^{(G)}(.)$ are defined in~\eqref{eq:global_reconstruction_loss} and 
$\hat{\mathbf{x}}$ is a reconstruction of $\mathbf{x}$ produced by the global IGD model. The $\ell^{(L)}_{r}\big (\mathbf{x}^{(L)}({\omega}) ,\hat{\mathbf{x}}^{(L)}({\omega}),\theta^*\big )$ in~\eqref{eq:localisation_loss^(G)lobal_local_MS-SSIM} is similarly defined using the local IGD model.
Thus, the anomaly localisation final map is a 
heatmap with high values representing regions that are likely to contain anomalies, as displayed in 'Global and Local IGD Models' section in the Supp. Material.
\section{Experiments}
\label{sec:datasets_evaluation}

%-------------------------------------------------------------------------
% {-.2cm}
\subsection{Datasets and Evaluation Metric}
% \textbf{Datasets:} We use four computer vision and two medical image datasets to evaluate our method. The computer vision datasets are MNIST~\cite{lecun2010mnist}, Fashion MNIST~\cite{fmnist}, CIFAR10~\cite{krizhevsky2014cifar} and MVTec AD~\cite{mvtecad}; and the medical image datasets are Hyper-Kvasir~\cite{borgli2020hyperkvasir} and LAG~\cite{li2019attention}. MNIST, Fashion MNIST and CIFAR10 have been widely used as benchmarks for image anomaly detection, and we follow the same experimental protocol as described in~\cite{dsvdd}. 
% We also tested our method on two publicly available medical datasets: Hyper-Kvasir~\cite{borgli2020hyperkvasir} and LAG~\cite{li2019attention} for polyp~\cite{tian2019one,pu2020computer,tian2021detecting} and glaucoma detection~\cite{tian2021constrained,tian2021selfsupervised,venkataramanan2019attention}, respectively. Please see the detailed description of the datasets in Supp. Material.
% \yuanhong{I moved the statistics of the dataset into the appendix.}

\textbf{Datasets:} We use four computer vision and two medical image datasets to evaluate our methods. The computer vision datasets are MNIST~\cite{lecun2010mnist}, Fashion MNIST~\cite{fmnist}, CIFAR10~\cite{krizhevsky2014cifar} and MVTec AD~\cite{mvtecad}; and the medical image datasets are Hyper-Kvasir~\cite{borgli2020hyperkvasir} and LAG~\cite{li2019attention}. MNIST, Fashion MNIST and CIFAR10 have been widely used as benchmarks for image anomaly detection, and we follow the same experimental protocol as described in~\cite{dsvdd}. 
CIFAR10 contains 60,000 images with 10 classes. MNIST and Fashion MNIST contain 70,000 images with 10 classes of handwritten digits and fashion products, respectively. 
MVTec AD~\cite{mvtecad} contains 5,354 high-resolution real-world images of 15 different industry object and textures. The normal class of MVTec AD is formed by 3,629 training and 467 testing images without defects. The anomalous class has more than 70 categories of defects (such as dents, structural fails, contamination, etc.) and contains 1,258 testing images. MVTec AD provides pixel-wise ground truth annotations for all anomalies in the testing images, allowing the evaluation of anomaly detection and localisation. 
We also tested our method on two publicly available medical datasets: Hyper-Kvasir~\cite{borgli2020hyperkvasir} and LAG~\cite{li2019attention} for polyp and glaucoma detection, respectively. For Hyper-Kvasir, we has 1,600 normal images without polyps in the training set and 500 in the testing set; and 1,000 abnormal images containing polyps in the testing set. For LAG, we have 2,343 normal images without glaucoma in the training set; and 800 normal images and 1,711 abnormal images with glaucoma for testing.

%-------------------------------------------------------------------------

\textbf{Evaluation:} For \emph{anomaly detection}, we assess performance with the area under the receiver operating characteristic curve (AUC) and classification accuracy. %~\cite{pangguansong2020}. 
On MNIST, Fashion MNIST and CIFAR10, we use the same protocol as other methods in Tab.~\ref{tab:auc_detection_mnist_cifar10_fmnist}, where training uses a single class as the normal data, with the nine remaining classes denoting as semantically anomalous samples, and inference relies on a non-augmented test image. 
%\gustavo{Training uses 5000 training images of the normal class and testing uses the 10000 images of the normal and anomalous classes.}
We report the mean AUC over the 10 classes for the above three data sets. On MVTec AD~\cite{bergmann2020uninformed,venkataramanan2019attention}, we evaluate anomaly detection with mean AUC and accuracy. Follow previous works~\cite{tian2021constrained,tian2021selfsupervised}, we evaluate the methods using AUC for the Hyper-Kvasir and LAG. 
For \emph{anomaly localisation}, we follow~\cite{venkataramanan2019attention} and compute the mean pixel-level AUC between the generated heatmap and the ground truth segmentation map for each anomalous image in the testing set of MVTec AD.

\bgroup
\def\arraystretch{1.1}
\begin{table}[t!]
\centering
\resizebox{0.98\linewidth}{!}{%
\begin{tabular}{@{}c|c|ccc@{}}
\toprule
Pretrain &Method                   &	MNIST	        &	CIFAR10	 & FMNIST        \\		\midrule\midrule
\multirow{16}{*}{Scratch}
&DAE~\cite{hadsell2006dimensionality}                         &	0.8766	        &	0.5358	 &  -       \\	
&VAE~\cite{vae}	                        &	0.9696	        &	0.5833     &  - 	\\
&KDE~\cite{bishop2006pattern}                         & 0.8140       &		0.6100	   &  -     \\
&OCSVM~\cite{oc-svm}	        &	0.9510	        &	0.5860     &  - 	\\	
&AnoGAN~\cite{anogan}	                    &	0.9127	        &	0.6179     &  - 	\\	
&DSVDD~\cite{dsvdd}	                    &	0.9480	        &	0.6481      &  -	\\	
&OCGAN~\cite{ocgan}	                    &	0.9750	        &	0.6566	    &  -    \\
&PixelCNN~\cite{van2016conditional}	                    & 0.6180	  	&	0.5510 	   &  -     \\
&CapsNet\textsubscript{PP}~\cite{li2020exploring}	                    &0.9770	   	&0.6120		 &  0.7650       \\
&CapsNet\textsubscript{RE}~\cite{li2020exploring}	                    &0.9250	   	& 0.5310		&  0.6790        \\
&ADGAN~\cite{ADGAN}	                    & 0.9680	   	& 0.6340	&  -	        \\
&LSA~\cite{lsa}	                    & 0.9750	   	& 0.6410		 &  0.8760       \\
&MemAE~\cite{gong2019memorizing}	                    &	0.9751      	&	0.6088	   &  -       \\	
&GradCon~\cite{gradcon}	                    & 0.9730	   	&	0.6640	    &  -      \\
&$\lambda$-VAE\textsubscript{u}~\cite{lamda-vae}	                    & 0.9820	   	& 0.7170	    &  0.8730     \\
&ULSLM~\cite{ulslm}	                    & 0.9490	   	& 0.7360		 &  -         \\ 
&SCADN~\cite{yan2021learning}       & 0.9771    & 0.6690 & -    \\
&\textbf{Ours}	            & \textbf{0.9869}		&	\textbf{0.7433}	 & \textbf{0.9201}
\\\hline
\multirow{3}{*}{ImageNet} &CAVGA-D\textsubscript{u}~\cite{venkataramanan2019attention} 	&	0.9860        &	0.7370	    &  0.8850      \\	
% \textbf{Ours - no KD}	            &	\textbf{?}	&	\textbf{?} \\
% &\textbf{Ours}	            &	\textbf{0.9912}	&	\textbf{0.8234}  &	\textbf{0.9332} \\
 &Student-Teacher~\cite{bergmann2020uninformed}	        &	\textbf{0.9935}	        &	0.8196	  &  -      \\
&\textbf{Ours}	            &	\textbf{0.9927}	&	\textbf{0.8368}	 & \textbf{0.9357} 
\\\hline
\multirow{3}{*}{SSL}
&Rot-Net~\cite{golan2018deep}	        &	-        &	0.8160	  &  0.9350     \\	
 &\citet{bergman2020classification}	        &	-        &	0.8820	  &  0.9410     \\
&\textbf{Ours}	            &	-	& 	\textbf{0.9125}	 & \textbf{0.9441}\\ 

\bottomrule
\end{tabular}}
\caption{\textbf{Anomaly detection:} mean AUC testing results on MNIST, CIFAR10 and Fashion MNIST. The results are split into 'Scratch' (without any pre-training), pretrained with 'ImageNet', and self-supervised learning ('SSL'). Bold numbers represent the best result (within 0.5\%) for each data set, discriminated by Scratch, SSL or ImageNet.}
\label{tab:auc_detection_mnist_cifar10_fmnist}
\end{table}

% {-.2cm}
\subsection{Implementation Details}

We implement our framework using Pytorch. The model was trained with Adam optimiser using a learning rate of 0.0001, weight decay of $10^{-6}$, batch size of 64 images, 256 epochs for all dataset. We defined the representation space produced by the encoder to have $Z=128$ dimensions.
Following~\cite{depthestimation2017}, we set $\rho=0.15$ to balance the contribution of MAE and MS-SSIM losses in~\eqref{eq:global_reconstruction_loss} and~\eqref{eq:localisation_map}. 
We set $\lambda_1 = \lambda_2 = 1$ in~\eqref{eq:loss} and  $\lambda_3=0.1$ in \eqref{eq:loss_AE}, based on cross validation experiments. We use Resnet18 and its reverse architecture as the encoder and decoder for both the global and local IGD models. 
When computing the accuracy of anomaly detection in MVTec AD, the threshold of the anomaly detection score $s(\mathbf{x})$ in~\eqref{eq:anomaly_score_image} (to classify an image as anomalous) is set to 0.5~\cite{venkataramanan2019attention}.
To enable a fair comparison between our method and previous approaches in the field~\cite{bergmann2020uninformed,venkataramanan2019attention,bergman2020classification,golan2018deep}, we pre-train the encoders for the global and local IGD models either with self-supervised learning (SSL)~\cite{chen2020simple}
or ImageNet knowledge distillation (KD)~\cite{bergmann2020uninformed,gou2020knowledge}. 
% Please see Supp. Material for more  information on the implementation. 
% \gustavo{I don't see anything in the appendix...}
For this SSL pre-training, we use the SGD optimiser with a learning rate of 0.01,  weight decay $10^{-1}$, batch size of 32, and 2,000 epochs. Once we obtain the pre-trained encoder with SSL, we remove the MLP layer and attach a linear layer to the backbone with fixed parameters. Note that this SSL is trained from scratch. 
In contrast to the vanilla self-supervised learning~\cite{chen2020simple} suggesting large batch size, we notice that a medium batch size yields significantly better performance for unsupervised anomaly detection. 
% KD

For the ImageNet KD pre-training, we minimise the $\ell2$ norm between the 512-dimensional feature vector output from encoder and an intermediate layer of the ImageNet pre-trained ResNet18 with the same 512-dimensional features.
For this ImageNet KD pre-training, we use the Adam optimiser with a learning rate of 0.0001,  weight decay $10^{-5}$, batch size of 64, and 50,000 iterations. Once we obtain the pre-trained encoder of KD, we fix the network parameters 
and attach a linear layer to reduce the dimensionality of the feature space to 128.

\begin{figure}[t]
    \centering
    \includegraphics[width=1\linewidth]{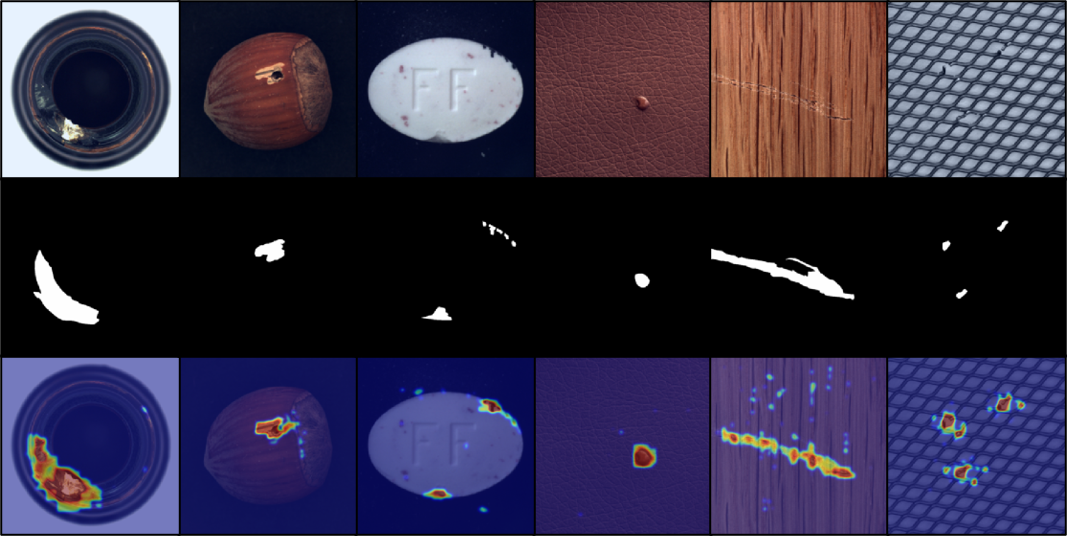}
    \caption{Qualitative results of our anomaly localisation results on the MVTec AD (red = high probability of anomaly). Top, middle and bottom rows show the testing images, ground-truth masks and predicted heatmaps, respectively. Please see additional results in the Supp. Material.} 
    \label{fig:mvtec_seg}
\end{figure}

% {-.2cm}
\subsection{Experiments on MNIST, Fashion MNIST and CIFAR10}
Table~\ref{tab:auc_detection_mnist_cifar10_fmnist} compares the unsupervised anomaly detection mean AUC testing results between our method and the current SOTA on MNIST, Fashion MNIST and CIFAR10.
The rows labelled as `Scratch' show results of models that were not pre-trained, and the ones with `SSL' display results from models using self-supervised learning method~\cite{golan2018deep,bergman2020classification}. The ones with `ImageNet' show results from models that use ImageNet KD pre-training~\cite{venkataramanan2019attention,bergmann2020uninformed}. 
Our proposed IGD 
outperforms current SOTA methods for the majority of pre-training methods on all three datasets.
Please see additional results in the Supp. material.

\subsection{Experiments on MVTec AD}

We report the results, based on SSL and ImageNet KD pre-trained models, for both anomaly detection (Tab.~\ref{tab:auc_detection_mvtec_short}) and localisation (Tab.~\ref{tab:MVTec-localisation-AUC}) on MVTec AD, which contains real-world images of industry objects and textures containing different types of anomalies. Following~\cite{venkataramanan2019attention} the score threshold is set to 0.5 for calculating the mean accuracy of anomaly detection.
For anomaly detection, our method produces the best accuracy (at least 2\% better than previous SOTA) and AUC (at least 5\% better than previous SOTA) results independently of the pre-training technique.
For anomaly localisation, we compare our method and the SOTA using the mean pixel-level AUC of all anomalous images in the testing set of MVTec AD. Notice that our method with ImageNet and SSL pre-training are better than the previous SOTA CAVGA-R\textsubscript{u}~\cite{venkataramanan2019attention} by 2\% and 4\%, respectively.
% It is also important to note that our architecture is lighter then CAVGA-R\textsubscript{u} given that it does not use an attention module and a visual discriminator. 
Fig.~\ref{fig:mvtec_seg} shows anomaly localisation results on MVTec AD images, where red regions in the heatmap indicate higher anomaly probability. From this results, we can see that our approach can localise anomalous regions of different sizes and structures from different object categories.
Please see additional results in the Supp. material.

\bgroup
\def\arraystretch{1.1}
\begin{table}[t]
\centering
\resizebox{0.92\linewidth}{!}{%
% \bgroup
% \def\arraystretch{1.5}
\begin{tabular}{@{}c|cc@{}}
\toprule 
Metric   & Method & Mean  \\ \hline \hline
       \multirow{10}{*}{Accuracy}   & AVID~\cite{sabokrou2018adversarially}              
         & 0.730  \\
         & AE\textsubscript{SSIM}~\cite{ae-ssim}   
          & 0.630  \\
         & DAE~\cite{hadsell2006dimensionality}   & 0.710  \\
         & AnoGAN~\cite{anogan} & 0.550  \\ 
         & $\lambda$-VAE\textsubscript{u}~\cite{lamda-vae}  & 0.770  \\
         & LSA~\cite{lsa}  & 0.730  \\
         & CAVGA-D\textsubscript{u}~\cite{venkataramanan2019attention}  & 0.780  \\
         & CAVGA-R\textsubscript{u}~\cite{venkataramanan2019attention}  & 0.820  \\ 
         & \textbf{Ours - ImageNet} & \textbf{0.840}  \\
         & \textbf{Ours - SSL} & \textbf{0.850}  \\ \midrule
 \multirow{7}{*}{AUC}        & AnoGAN~\cite{anogan} & 0.503 \\
         & GANomaly~\cite{akcay2018ganomaly}& 0.782 \\
         & Skip-GANomaly~\cite{akccay2019skip}   & 0.805 \\
         & SCADN~\cite{yan2021learning} & 0.818 \\
     & U-Net~\cite{u-net}   & 0.819 \\
         & DAGAN~\cite{DAGAN}  & 0.873 \\
         & \textbf{Ours - ImageNet} & {\textbf{0.926}} \\
         & \textbf{Ours - SSL}  & {\textbf{0.934}} \\ \bottomrule
\end{tabular}%
}
\caption{
\textbf{Anomaly detection}: mean testing accuracy and AUC on MVTec AD produced by the SOTA and our IGD.}
\label{tab:auc_detection_mvtec_short}
\end{table}

\bgroup
\def\arraystretch{1.1}
\begin{table}[htp]
\centering
\small
\resizebox{0.82\linewidth}{!}{%
\begin{tabular}{@{}cc@{}}
\toprule
Method	                    &	MVTec AD      	\\	\midrule\midrule    %\hline\hline
DAE~\cite{hadsell2006dimensionality}	    &	0.82        	\\	
AE\textsubscript{SSIM}~\cite{ae-ssim}	    &	0.87	        \\	
AVID~\cite{sabokrou2018adversarially}	    & 0.78		        \\
SCADN~\cite{yan2021learning}                & 0.75  \\
LSA~\cite{lsa}	    &	0.79	        \\ 
$\lambda$-VAE\textsubscript{u}~\cite{lamda-vae}	    &	0.86	        \\
AnoGAN~\cite{anogan}	                    &	0.74	        \\	ADVAE~\cite{ADVAE}	                    &	0.86	        \\
CAVGA-D\textsubscript{u}~\cite{venkataramanan2019attention}	&	0.85	        \\	
CAVGA-R\textsubscript{u}~\cite{venkataramanan2019attention}	&	0.89	        \\	
\textbf{Ours - ImageNet}	            &	{\textbf{0.91}}	\\
\textbf{Ours - SSL} & {\textbf{0.93}}	\\\bottomrule
\end{tabular}}
% {.3in}
\caption{\textbf{Anomaly localisation:} mean pixel-level AUC testing results on the anomalous images of MVTec AD. }
\label{tab:MVTec-localisation-AUC}
\end{table}

\subsection{Experiments on Medical Datasets}

To show that our method can generalise to other domains, we evaluate our approach on two public medical datasets - Hyper-Kvasir for polyp detection and LAG for glaucoma detection. As shown in Tab.~\ref{tab:medical_auc}, our
SSL and ImageNet based results achieve the best AUC results on both datasets. Our methods surpass the recent proposed CAVGA-R\textsubscript{u}~\cite{venkataramanan2019attention} on both datasets by a minimum 0.9\% and maximum 3.8\%. Also, our model performs better compared to the anomaly detector specifically designed for medical data, such as f-anogan~\cite{f-anogan} and ADGAN~\cite{liu2019photoshopping}.
We show qualitative polyp segmentation results in the Supp. Material.
The abnormalities in medical data (i.e., colon polyps, glaucoma) are significantly different than the popular image benchmarks and MVTec AD in terms of appearance and structural anomalies, suggesting that our model works in disparate domains.
Please see more results in the Supp. material.

\bgroup
\def\arraystretch{1.1}
\begin{table}
\centering
\small
\resizebox{0.92\linewidth}{!}{%
\begin{tabular}{@{}ccc@{}}
\toprule 
Methods         & Hyper-Kvasir  & LAG  \\ \hline\hline
DAE~\cite{masci2011stacked}             & 0.705      & 0.651   \\
CAM~\cite{cam}                &           -      &   0.663      \\
GBP~\cite{springenberg2014striving}                &   -             & 0.787         \\
SmoothGrad~\cite{smilkov2017smoothgrad}                &  -               & 0.795  \\
OCGAN~\cite{perera2019ocgan}           & 0.813     &  0.737 \\
F-anoGAN~\cite{f-anogan}        & 0.907       & 0.778  \\
ADGAN~\cite{liu2019photoshopping}           & 0.913      &   0.752    \\
CAVGA-$R_{u}$~\cite{venkataramanan2019attention}     & 0.928    & 0.819       \\
\textbf{Ours - ImageNet}      & {\textbf{0.931}}        & {\textbf{0.838}}   \\
\textbf{Ours - SSL}    & {\textbf{0.937}}        & {\textbf{0.857}}   \\\bottomrule
\end{tabular}
}
\caption{\textbf{Anomaly detection:} AUC testing results on two medical datasets: Hyper-Kvasir and LAG.}
\label{tab:medical_auc}
\end{table}

% \begin{table}[h]
% \centering
% \scalebox{0.8}{

% {-.2cm}
\subsection{Ablation Study}
To investigate the effectiveness of each component of our method, we show the mean AUC results of our method with different proposed variants in Tab.~\ref{tab:ablation}. Note that all results are based on the initialisation of knowledge distillation from ImageNet. For standard anomaly detection settings (AUC - Full), each proposed component of our IGD improves performance by a minimum 1.7\% and maximum 11.6\% mean AUC.  Tab.~\ref{tab:ablation} also shows the effectiveness of each component when trained with small (20\% of full training data) or anomaly contaminated (10\% of contamination rate) training sets, where our proposed Gaussian anomaly classifier (GAC) significantly improves over the REC (i.e., MS-SSIM+MAE losses) baseline by 13\% and 10.4\% mean AUC. The proposed adversarial interpolation regularisation (INTER) further improves the AUC by 3.7\% and 3.1\%. 

\begin{table}[t]
\centering
\scalebox{0.7}{
% \resizebox{\linewidth}{!}{%
\begin{tabular}{cccc|ccc}
\toprule%\hline
  MSE	&	REC 	&	GAC	&	INTER	        &	AUC - Full  & AUC - ST &	AUC - AC	\\ \midrule\midrule%
\checkmark	&		&	&		                    &   0.615       & 0.552    &	0.565       \\
	&	\checkmark &			&		            &	0.731       & 0.655    &	0.677       \\
&	\checkmark	& 		\checkmark	&		        &	0.819       & 0.785    &	0.781       \\
	&	\checkmark	& 	\checkmark	&	\checkmark	&	0.836       & 0.822	   &    0.812       \\ \hline\bottomrule
\end{tabular}%
}
\caption{Ablation study of our method on CIFAR10 using anomaly detection mean testing AUC w.r.t standard OCC setup (AUC - Full), small training set containing 20\% of training data (AUC - ST), and anomaly contaminated training set with 10\% contamination (i.e., 10\% of the anomalous samples are removed from the testing set and inserted into the training set) (AUC - AC).
MSE denotes the baseline deep autoencoder with MSE loss, REC denotes the baseline deep autoencoder with MS-SSIM + MAE losses, GAC denotes our proposed Gaussian anomaly classifier, INTER represents our interpolation regularisation. The encoder of all above methods are initialised based on the knowledge distillation from ImageNet.  } 
\label{tab:ablation}
\end{table}

% {-.2cm}
\subsection{Experiments on Small/Contaminated Training Sets}

\begin{table}[t]
\centering
\begin{center}
\resizebox{0.9\linewidth}{!}{\centering{%
\begin{tabular}{@{}c| cccc @{}}
\toprule
Dataset & Train Size &	DSVDD	& DSVDD+REC	&\textbf{IGD (Ours)}	\\ \hline	 \hline			
\multirow{3}{*}{CIFAR10}
& 20\%	&	0.7064	&	0.7462	&	0.8219	\\		
& 60\%	&	0.7367	&	0.7807	&	0.8298	\\ 					
& 100\%	&	0.7612	&	0.7950	&	0.8365	\\ 
\midrule 
\multirow{3}{*}{MVTec}
& 20\%	&	0.7994  &	0.7291  &	0.9043  \\
& 60\%	&	0.8467  &	0.7737  &	0.9246  \\
& 100\% &	0.8579  &	0.7826  &	0.9260  \\ 
\bottomrule \bottomrule
\end{tabular}}}
\end{center}
\caption{Mean testing AUCs on CIFAR10 and MVTec with small training sets, where REC=MS-SSIM+MAE losses.}
\label{tab:sample_efficiency}
\end{table}

To show the improved robustness of our approach to small training sets on CIFAR10 and MVTec, we compare the performance of DSVDD, DSVDD+REC (i.e., DSVDD combined with our reconstruction loss), and our proposed IGD, using less normal data in the training sets in Tab.~\ref{tab:sample_efficiency}.  In particular, we randomly sub-sample 20\%, 60\%, and 100\% of the original training sets of CIFAR10 and MVTec AD, to form a smaller training set. The results indicate that IGD achieves comparable performance under significantly less training data, while the performance of DSVDD and DSVDD+REC deteriorate dramatically when the number of training samples decreases.  
This result shows that IGD has better robustness than DSVDD and DSVDD+REC to small training sets. 

% \subsection{Experiments on Anomaly Contamination}
\begin{table}[t]
\centering
\begin{center}
\resizebox{0.9\linewidth}{!}{\centering{%
\begin{tabular}{@{}c| cccc @{}}
\toprule
Dataset  & Noise Ratio	&	DSVDD	& DSVDD+REC	& \textbf{IGD (Ours)}	\\ \hline	 \hline	
\multirow{3}{*}{CIFAR10}
& 1\%	 &	0.7502	&	0.7694	&	0.8252	\\												
& 5\%	 &	0.7124	&	0.7448	&	0.8193	\\												
& 10\%	 &	0.6717	&	0.7073	&	0.8122	\\ 
\midrule 
\multirow{3}{*}{MVTec}
& 1\%	 &	0.8523	&	0.7873	&   0.9363	\\
& 5\%	 &	0.8391	&	0.7733 	&	0.9319	\\			
& 10\%   &	0.8175	&	0.7687	&	0.9363	\\ 
\bottomrule \bottomrule
\end{tabular}}}
\end{center}
\caption{Mean testing AUCs on CIFAR10 and MVTec with different contamination noise rates. REC defined in Tab.~\ref{tab:sample_efficiency}.}
% {-.18in}
\label{tab:noise_conta}
\end{table}

To show the improved robustness of our approach contaminated training sets, in Tab.~\ref{tab:noise_conta},  we compare the performance of DSVDD, DSVDD+REC, and our IGD, using training sets corrupted with anomalous samples (this contamination facilitates overfitting).
In particular, we re-organise the original training and test data of CIFAR10 and MVTec AD by randomly sampling 1\%, 5\% and 10\% of anomalies from the test data to inject into the training data. With different rates of anomaly contamination, the maximum fluctuation of our IGD is 1.3\% on CIFAR10 and 0.44\% on MVTec AD. While the competing method DSVDD shows a much larger maximum fluctuation of 7.8\% and 3.5\% mean AUC, on CIFAR10 and MVTec AD, respectively.   The results show the substantially better robustness of IGD over DSVDD and DSVDD+REC for the anomaly-contaminated training data. 

\section{Discussion}
We do not compare some of the SOTA works~\cite{reiss2021panda,sohn2020learning,tack2020csi} in Table \ref{tab:auc_detection_mnist_cifar10_fmnist}, \ref{tab:auc_detection_mvtec_short}, and \ref{tab:MVTec-localisation-AUC} due to unfair comparison. 
In particular, the comparison with PANDA~\cite{reiss2021panda} is not fair because it uses a WideResNet50 $\times$ 2 for MVTec and ResNet152 for CIFAR, both being much larger backbones than our ResNet18. 
Regarding CSI~\cite{tack2020csi}, it has much slower inference (because of the $40\times$ data augmentation of test images) and more complex training that needs a coreset and large batch size of 512 for pre-training, which challenges its use for problems with small training sets or high-resolution images.
For both CSI and DROC~\cite{sohn2020learning}, their gains are mostly from the SSL pre-training. To show that point for CSI, we use our training approach to fine-tune a pre-trained CSI model and obtain 94.6\% AUC on CIFAR10, which is higher than CSI (94.3\% AUC).
% surpasses CSI's (94.3). 
Also, for the vanilla SSL pre-training reported in DROC paper, their performance reduces from 92.5\% to 89.0\% AUC on CIFAR10, and from 86.5\% to 80.2\% AUC on MVTec. 
Note that all above results are collected from their published papers unless stated otherwise. 

Furthermore, on MVTec, our approach obtains (93.4\% AUC), which is much better than CSI (63.6\% AUC from Tab.2 of~\cite{reiss2021mean}) and PANDA (86.5\%).
For anomaly localisation on MVTec, our 93\% AUC is better than DROC (90\%) and worse than PANDA (96\%).
On high-resolution image datasets (e.g., Hyper-Kvasir), our approach (93.7\% AUC) is better than CSI (trained by us) that reaches 91.6\% AUC. 
Other important results shown by our paper, but missed by CSI, PANDA and DROC, are the ones with small training sets and contaminated training sets, which are new and important benchmarks for real-world industrial applications and early detection of medical diseases.

\section{Conclusion}
% In this paper, we presented a new EM based unsupervised anomaly detection and localisation method and a new criterion to detect and localise structural and non-structural multi-scale anomalies. We use E-step to estimate the parameters of the distribution of normal images, and the M-step containing an optimisation of Gaussian classifier with the constraints of multi-scale image reconstruction and adversarial image representation interpolation.

In this paper, we presented a new
OCC model, called interpolated Gaussian descriptor (IGD), to perform unsupervised anomaly detection and segmentation.
IGD learns a one-class Gaussian anomaly classifier trained with adversarially interpolated training samples to enable an effective normality description based on representative normal samples rather than fringe or anomalous samples.
The optimisation of IGD is formulated as an EM algorithm, which we show to be theoretically correct and to converge to a stationary solution under certain conditions.
% unsupervised anomaly detection and localisation method which learns a distribution of normal images that generalises well to all classes of normal images, a new criterion to detect and localise structural and non-structural multi-scale anomalies, and a new training scheme that based on EM optimisation.
% We use E-step to estimate the parameters of the distribution of normal images, and the M-step containing an optimisation of Gaussian classifier with the constraints of multi-scale image reconstruction and adversarial image representation interpolation.
% Our learned distribution of normal images is based on the interpolated Gaussian descriptor (IGD) that provides a robust modelling of the normal image distribution with adversarially interpolated descriptors to regularise and facilitate the training of the proposed Gaussian anomaly classifier.  
% The proposed reconstruction loss can find structural and non-structural anomalies of varying sizes.
% The IGD and reconstruction losses are then jointly optimised to produce 
%The IGD is further extend to a global and a local model that 
To our knowledge, IGD is the first method that is able to achieve the best performance across diverse application datasets, including MNIST, CIFAR10, Fashion MNIST, MVTec AD, and two large scale medical datasets, in terms of anomaly detection and localisation. 
We also show that IGD is more robust than DSVDD and an image-reconstruction contrained DSVDD in problems with small or contaminated training sets.
We plan to study the use of Gaussian anomaly classifier in the pixel-wise localisation of anomalies and to investigate new self-supervised learning approaches specifically designed for anomaly detection.

\clearpage
\small{\bibliography{aaai22.bib}}

\clearpage
\newpage

\beginsupplement
\setcounter{section}{0}
\setcounter{equation}{0}
\setcounter{figure}{0}
\setcounter{table}{0}
\setcounter{page}{1}
\makeatletter
\renewcommand{\theequation}{S\arabic{equation}}
\renewcommand{\thefigure}{S\arabic{figure}}
\newpage

\section{Datasets}
CIFAR10 contains 60,000 images with 10 classes. MNIST and Fashion MNIST contain 70,000 images with 10 classes of handwritten digits and fashion products, respectively. 
MVTec AD~\cite{mvtecad} contains 5,354 high-resolution real-world images of 15 different industry object and textures. The normal class of MVTec AD is formed by 3,629 training and 467 testing images without defects. The anomalous class has more than 70 categories of defects
(such as dents, structural fails, contamination, etc.) and contains 1,258 testing images. MVTec AD provides pixel-wise ground truth annotations for all anomalies in the testing images, allowing the evaluation of anomaly detection and localisation. 
Hyper-Kvasir has 1,600 normal images without polyps in the training set and 500 in the testing set; and 1,000 abnormal images containing polyps in the testing set. For LAG, we have 2,343 normal images without glaucoma in the training set; and 800 normal images and 1,711 abnormal images with glaucoma for testing.

\section{Global and Local IGD Models}

Figure~\ref{fig:multi-test} shows an example of a multi-scale structural and non-structural anomaly localisation result for an MVTec AD image, using both the local and global IGD models.

\begin{figure}[htp]
    \centering
    \includegraphics[width=\linewidth]{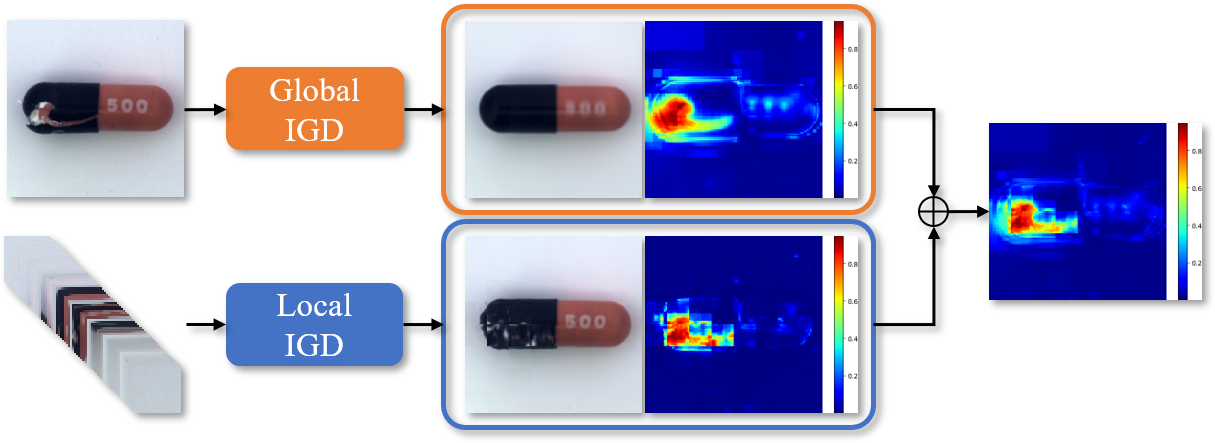}
   \caption{Example of the multi-scale structural and non-structural anomaly localisation result for an MVTec AD~\cite{mvtecad} image, using both the local and global IGD models. The global model tends to produce smooth results but with some mistakes, while the local model produces jagged results, but without the global mistakes, so by combining the two results, we obtain a smooth and correct anomaly heatmap.  %\yu{change figure to Global IGD and Local IGD}
   %\yuanhong{Can we move this figure to appendix?} \yu{If we don't have space,  I think we can. }
   }
    \label{fig:multi-test}
\end{figure}

\section{Multi-scale Structure Similarity Index (MS-SSIM) Score}
The MS-SSIM loss uses the MS-SSIM global score, defined as 
\begin{equation}
\begin{split}
    m^{(G)}(\mathbf{x}({\omega}),&\hat{\mathbf{x}}({\omega})) =  [l_M(\mathbf{x}({\omega}),\hat{\mathbf{x}}({\omega}))]^{\alpha_M} \times \\ & \prod_{m=1}^{m^{(G)}}
    [c_m(\mathbf{x}({\omega}),\hat{\mathbf{x}}({\omega}))]^{\beta_m}[s_m(\mathbf{x}({\omega}),\hat{\mathbf{x}}({\omega}))]^{\gamma_m},
    \label{eq:MS-SSIM}
\end{split}    
\end{equation}
where $\mathbf{x}({\omega})$ denotes an image patch centred at $\omega \in \Omega$ of size $11 \times 11 \times 3$,
\begin{equation}
    l_M(\mathbf{x}({\omega}),\hat{\mathbf{x}}({\omega})) =  \frac{2\mu_{\mathbf{x}({\omega})}\mu_{\hat{\mathbf{x}}({\omega})}+C_1}{\mu_{\mathbf{x}({\omega})}^2 + \mu_{\hat{\mathbf{x}}({\omega})}^2 + C_1},
\end{equation}
\begin{equation}
 c_m(\mathbf{x}({\omega}),\hat{\mathbf{x}}({\omega})) =  \frac{2\sigma_{\mathbf{x}({\omega})}\sigma_{\hat{\mathbf{x}}({\omega})}+C_2}{\sigma_{\mathbf{x}({\omega})}^2 + \sigma_{\hat{\mathbf{x}}({\omega})}^2 + C_2},
 \end{equation}
 \begin{equation}
s_m(\mathbf{x}({\omega}),\hat{\mathbf{x}}({\omega})) = \frac{\sigma_{\mathbf{x}({\omega})\hat{\mathbf{x}}({\omega})}+C_3}{\sigma_{\mathbf{x}({\omega})}
\sigma_{\hat{\mathbf{x}}({\omega})} + C_3},   
 \end{equation}
 with $C_1,C_2,C_3$ representing pre-defined constants, $\mu_{\mathbf{x}({\omega})}$ denoting the mean intensities of $\mathbf{x}({\omega})$, $\sigma^2_{\mathbf{x}({\omega})}$
the variance of $\mathbf{x}({\omega})$, and
$\sigma_{\mathbf{x}({\omega})\hat{\mathbf{x}}({\omega})}$ the covariance of $\mathbf{x}({\omega})$ and $\hat{\mathbf{x}}({\omega})$. 
In~\eqref{eq:MS-SSIM}, 
$m^{(G)}=5$ denotes the number of scales,  $\beta_{1}=\gamma_{1}=0.0448$, $\beta_{2}=\gamma_{2}=0.2856$, $\beta_{3}=\gamma_{3}=0.3001$, $\beta_{4}=\gamma_{4}=0.2363$, $\alpha_{5}=\beta_{5}=\gamma_{5}=0.1333$~\cite{MS-SSIM}. %\gustavo{What about the $C_1,C_2,C_3$ values?} 
We follow $C_i=\left(K_iL\right)^2$ (for $i\in\{1,2,3\}$) according to~\cite{SSIM} and define $L=4.7579$ as the pixel range with $K_1=0.01$, $K_2=0.03$ and $C_3=C_2/2$.

The local score $m^{(L)}(\mathbf{x}^{(L)}({\omega}),\hat{\mathbf{x}}^{(L)}({\omega}))$ is defined in the same way as in~\eqref{eq:MS-SSIM}, where $\mathbf{x}^{(L)}({\omega})$ is an image patch centred at $\omega \in \Omega$ of size $3 \times 3 \times 3$,
$m^{(L)}=4$ scales with weights $\beta_{1}=\gamma_{1}=0.0516$, $\beta_{2}=\gamma_{2}=0.3295$, $\beta_{3}=\gamma_{3}=0.3463$, $\alpha_{4}=\beta_{4}=\gamma_{4}=0.2726$ modified based on the original proportion for $m^{(G)}=5$.

% \section{Details of the encoder, decoder, critic}
% =======================================================================
\section{Implementation Details}

For this SSL pre-training, we use the SGD optimiser with a learning rate of 0.01,  weight decay $10^{-1}$, batch size of 32, and 2,000 epochs. Once we obtain the pre-trained encoder with SSL, we remove the MLP layer and attach a linear layer to the backbone with fixed parameters. Note that this SSL is trained from scratch. 
In contrast to the vanilla self-supervised learning~\cite{chen2020simple} suggesting large batch size, we notice that a medium batch size yields significantly better performance for unsupervised anomaly detection. 
% KD

For the ImageNet KD pre-training, we minimise the $\ell2$ norm between the 512-dimensional feature vector output from encoder and an intermediate layer of the ImageNet pre-trained ResNet18~\cite{resnet} with the same 512-dimensional features.
For this ImageNet KD pre-training, we use the Adam optimiser with a learning rate of 0.0001,  weight decay $10^{-5}$, batch size of 64, and 50,000 iterations. Once we obtain the pre-trained encoder of KD, we fix the network parameters 
and attach a linear layer to reduce the dimensionality of the feature space to 128.
% =======================================================================

% \section{Visualisation of the Distribution of Normal Training Samples}
\section{Visualisation of the Distribution of Testing Samples}

Figure~\ref{fig:mvtec_tsne} shows the distribution of testing samples in the representation space, using the t-SNE visualisation, for DSVDD~\cite{dsvdd}, Gaussian anomaly classifier (GAC), and our IGD. Notice that the normal samples seem to be more compactly represented with fewer anomalous samples appearing inside the normal cluster. This suggests that IGD has a superior normality description, compared with DSVDD and GAC.

\begin{figure}[H]
  \centering
    \includegraphics[width=0.32\linewidth]{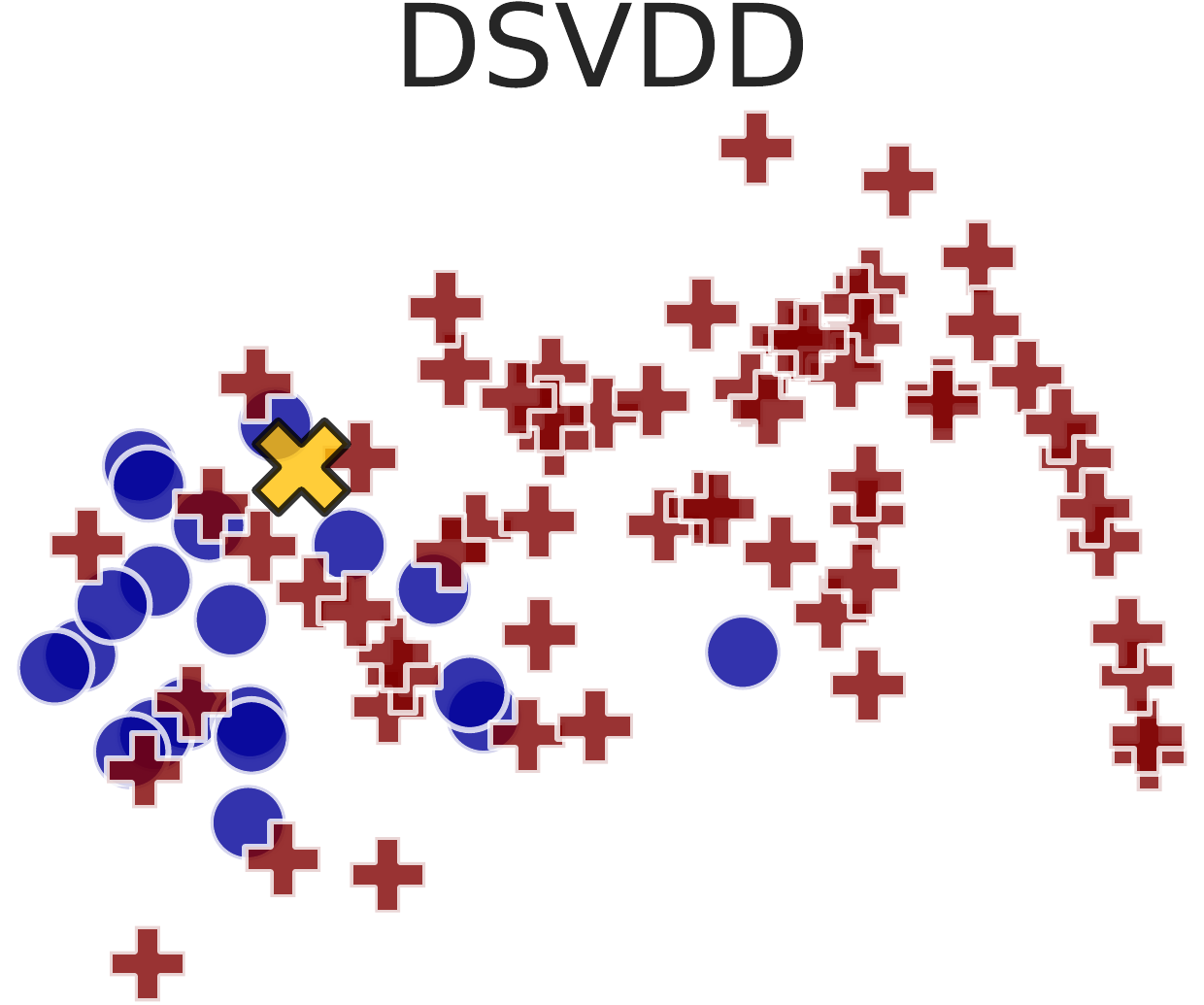}
    \includegraphics[width=0.32\linewidth]{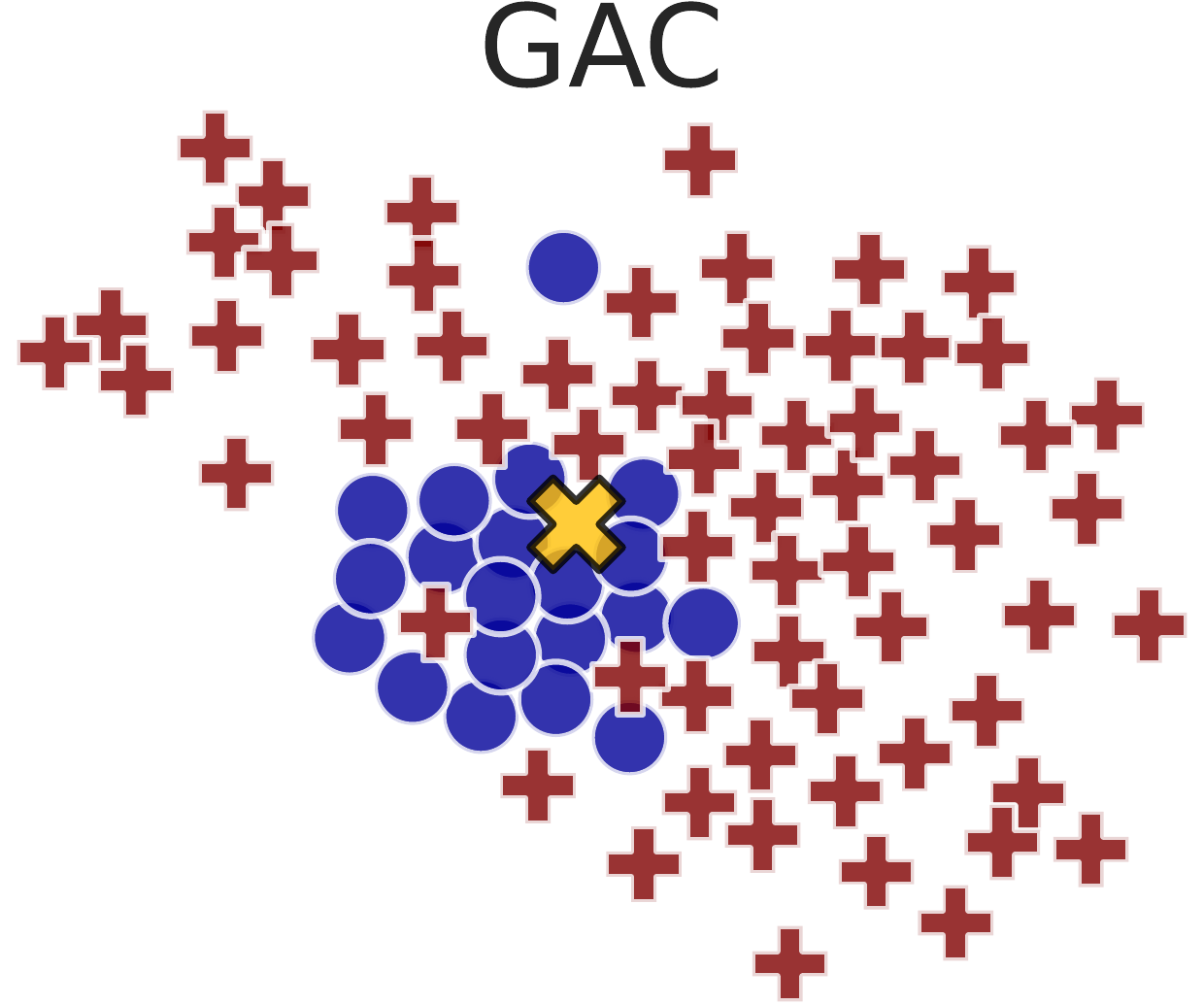}
    \includegraphics[width=0.32\linewidth]{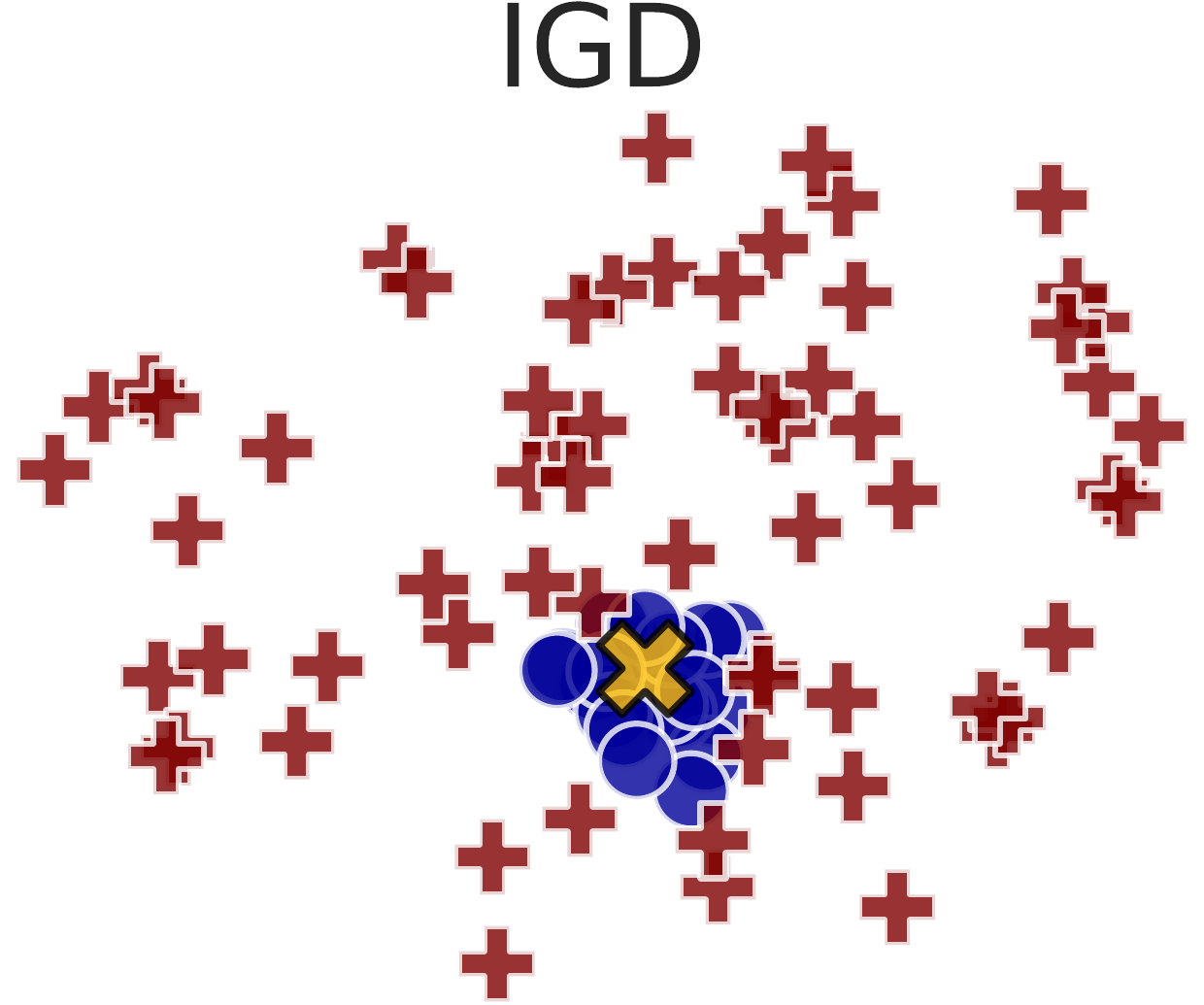}
    \includegraphics[width=0.5\linewidth]{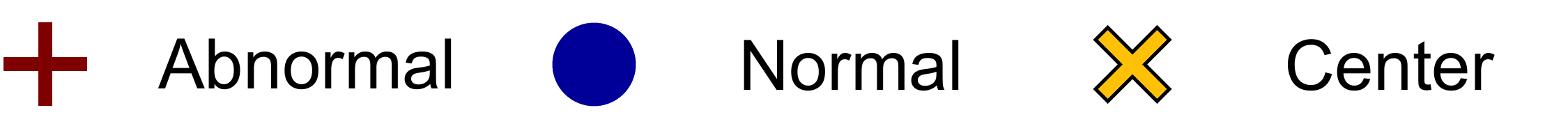}
    \caption{
    t-sne visualisation from MVTec (class bottle). 
    } 
    \label{fig:mvtec_tsne}
\end{figure}
\vspace{-.33cm}

\bgroup
\def\arraystretch{1.4}
\begin{table*}[h]
\centering
\resizebox{1\linewidth}{!}{%
% \bgroup
% \def\arraystretch{1.5}
\begin{tabular}{@{}cccccccccccccccccc@{}}
\toprule 
Metric   & Method                                                     & Bottle & Hazelnut & Capsule & Metal Nut & Leather & Pill  & Wood  & Carpet & Tile  & Grid  & Cable & Transistor & Toothbrush & Screw & Zipper & Mean  \\ \hline \hline
       \multirow{10}{*}{Accuracy}   & AVID~\cite{sabokrou2018adversarially}                      
         & 0.85   & 0.86     & 0.85    & 0.63      & 0.58    & 0.86  & 0.83  & 0.70    & 0.66  & 0.59  & 0.64  & 0.58       & 0.73       & 0.66  & 0.84   & 0.73  \\
         & AE\textsubscript{SSIM~\cite{ae-ssim}}                       
         & 0.88   & 0.54     & 0.61    & 0.54      & 0.46    & 0.60   & 0.83  & 0.67   & 0.52  & 0.69  & 0.61  & 0.52       & 0.74       & 0.51  & 0.80    & 0.63  \\
         & DAE~\cite{hadsell2006dimensionality}                       
         & 0.80    & 0.88     & 0.62    & 0.73      & 0.44    & 0.62  & 0.74  & 0.50    & 0.77  & 0.78  & 0.56  & 0.71       & \textbf{0.98}       & 0.69  & 0.80    & 0.71  \\
         & AnoGAN~\cite{anogan}                                       
         & 0.69   & 0.50      & 0.58    & 0.50       & 0.52    & 0.62  & 0.68  & 0.49   & 0.51  & 0.51  & 0.53  & 0.67       & 0.57       & 0.35  & 0.59   & 0.55  \\ 
         & $\lambda$-VAE\textsubscript{u~\cite{lamda-vae}}             
         & 0.86   & 0.74     & \textbf{0.86}    & \textbf{0.78}      & 0.71    & 0.80   & 0.89  & 0.67   & 0.81  & 0.83  & 0.56  & 0.70        & 0.89       & 0.71  & 0.67   & 0.77  \\
         & LSA~\cite{lsa}                                             
         & 0.86   & 0.80      & 0.71    & 0.67      & 0.70     & 0.85  & 0.75  & \textbf{0.74}   & 0.70   & 0.54  & 0.61  & 0.50        & 0.89       & 0.75  & \textbf{0.88}   & 0.73  \\
         & CAVGA-D\textsubscript{u~\cite{venkataramanan2019attention}} 
         & 0.89   & 0.84     & 0.83    & 0.67      & 0.71    & \textbf{0.88}  & 0.85  & 0.73   & 0.70   & 0.75  & 0.63  & 0.73       & 0.91       & \textbf{0.77}  & 0.87   & 0.78  \\
         & CAVGA-R\textsubscript{u~\cite{venkataramanan2019attention}} 
         & \textbf{0.91}   & \textbf{0.87}     & \textbf{0.87}    & 0.71      & 0.75    & \textbf{0.91}  & 0.88  & \textbf{0.78}   & 0.72  & 0.78  & 0.67  & 0.75       & \textbf{0.97}       & \textbf{0.78}  & \textbf{0.94}   & 0.82  \\ 
         & \textbf{Ours - ImageNet}                                                  
         & {\textbf{0.95}}   & \textbf{0.93}     & 0.80     & \textbf{0.82}      & \textbf{0.87}    & 0.77  & \textbf{0.94}  & 0.69   & \textbf{0.90}   & \textbf{0.92}  & \textbf{0.73}  & \textbf{0.88}       & \textbf{0.98}       & 0.58  & 0.85   & \textbf{0.84}  \\
         & \textbf{Ours - SSL}                                                 
         & {\textbf{0.95}}   & \textbf{0.93}     & 0.81    & \textbf{0.82}      & \textbf{0.90}     & 0.74  & \textbf{0.89}  & 0.71   & \textbf{0.94}  & \textbf{0.90}   & \textbf{0.79}  & \textbf{0.85}       & \textbf{0.98}       & 0.67  & \textbf{0.88}   & \textbf{0.85}  \\ \midrule
 \multirow{7}{*}{AUC}        & AnoGAN~\cite{anogan}                                       
         & 0.800    & 0.259    & 0.442   & 0.284     & 0.451   & 0.711 & 0.567 & 0.337  & 0.401 & 0.871 & 0.477 & 0.692      & 0.439      & 0.100   & 0.715  & 0.503 \\
         & GANomaly~\cite{akcay2018ganomaly}                          
         & 0.794  & 0.874    & 0.721   & 0.694     & 0.808   & 0.671 & 0.920  & 0.821  & 0.720  & 0.743 & 0.711 & 0.808      & 0.700        & {\textbf{1.000}}     & 0.744  & 0.782 \\
         & Skip-GANomaly~\cite{akccay2019skip}                        & 0.937  & 0.906    & 0.718   & 0.790      & 0.908   & 0.758 & 0.919 & 0.795  & 0.850  & 0.657 & 0.674 & 0.814      & 0.689      & {\textbf{1.000}}    & 0.663  & 0.805 \\
     & U-Net~\cite{u-net}                                         & 0.863  & 0.996    & 0.673   & 0.676     & 0.870    & 0.781 & 0.958 & 0.774  & 0.964 & 0.857 & 0.636 & 0.674      & 0.811      & {\textbf{1.000}}     & 0.750   & 0.819 \\
         & DAGAN~\cite{DAGAN}                                         & {\textbf{0.983}}  & {\textbf{1.000}}        & 0.687   & 0.815     & {\textbf{0.944}}   & 0.768 & {\textbf{0.979}} & {\textbf{0.903}}  & 0.961 & 0.867 & 0.665 & 0.794      & {\textbf{0.950}}       & {\textbf{1.000}}     & 0.781  & 0.873 \\
         & SCADN~\cite{yan2021learning} & 0.957 & 0.856 & 0.765 & 0.504 & 0.983 & 0.833 & 0.659 & 0.624 & 0.814 & 0.831 & 0.792 & 0.981 & 0.863 & 0.968 & 0.846 & 0.818\\
         & \textbf{Ours - ImageNet}                                                  & {\textbf{1.000}}      & 0.986    & {\textbf{0.907}}   & {\textbf{0.886}}     & 0.922   & {\textbf{0.870}}  & {\textbf{0.982}} & {\textbf{0.828}}  & {\textbf{0.979}} & {\textbf{0.979}} & {\textbf{0.856}} & {\textbf{0.909}}     & {\textbf{0.997}}      & 0.815 & {\textbf{0.969 }} & {\textbf{0.926}} \\
         & \textbf{Ours - SSL}                                                 & {\textbf{1.000}}     & {\textbf{0.997}}    & {\textbf{0.915}}   & {\textbf{0.913}}     & {\textbf{0.958}}   & {\textbf{0.873}} & 0.946 & {\textbf{0.828 }} & {\textbf{0.991}} & {\textbf{0.978}} & {\textbf{0.906}} &{\textbf{0.906 }}     & {\textbf{0.997}}     & {\textbf{0.825}} & {\textbf{0.970 }}  & {\textbf{0.934}} \\ \bottomrule
\end{tabular}%
}
\caption{
\textbf{Anomaly detection}: mean testing accuracy and AUC on MVTec AD produced by the SOTA and our method.
}
\label{tab:auc_class_mvtec}
\end{table*}

\section{Correctness Proof}
\label{sec:correctness_analysis_proof}

\begin{restatable}[]{lemma}{Correctness}
    \label{thm:correctness}
    Assuming that the maximisation of the constrained $\ell_{ELBO}$ produces $\theta$ that makes\\ 
    \scalebox{0.87}{
    $\mathbb{E}_{q(\omega)}[\log p_{\theta}(y=0,\omega|\mathbf{x},\mathcal{P}_{\mathcal{X}})] \ge 
    \mathbb{E}_{q(\omega)}[\log p_{\theta^{old}}(y=0,\omega|\mathbf{x},\mathcal{P}_{\mathcal{X}})]$,} \\we have that 
    \scalebox{0.87}{$\left(\log p_{\theta}(y=0|\mathbf{x},\mathcal{P}_{\mathcal{X}}) - \log p_{\theta^{old}}(y=0|\mathbf{x},\mathcal{P}_{\mathcal{X}})\right)$} 
    is lower bounded by\\
    \scalebox{0.8}{
    $\left(\mathbb{E}_{q(\omega)}[\log p_{\theta}(y=0,\omega|\mathbf{x},\mathcal{P}_{\mathcal{X}})] - 
    \mathbb{E}_{q(\omega)}[\log p_{\theta^{old}}(y=0,\omega|\mathbf{x},\mathcal{P}_{\mathcal{X}})]\right) \ge 0$,} \\
    with $q(\omega)=p_{\theta^{old}}(\omega|\mathcal{P}_{\mathcal{X}})$.
 \end{restatable}

\begin{proof}
We follow the proof for Theorem 1 in~\citep{dempster1977maximum}. From the main paper, we have
\begin{equation}
    \begin{split}
         \log & p_{\theta}(y=0|\mathbf{x},\mathcal{P}_{\mathcal{X}})=\\
         &\ell_{ELBO}(q,\theta)+KL[q(\omega)||p_{\theta}(\omega|\mathcal{P}_{\mathcal{X}})], 
    \end{split}
\end{equation}
where $q(\omega)=p_{\theta^{old}}(\omega|\mathcal{P}_{\mathcal{X}})$. Subtracting 
$\log p_{\theta}(y=0|\mathbf{x},\mathcal{P}_{\mathcal{X}})$ and $\log p_{\theta^{old}}(y=0|\mathbf{x}\mathcal{P}_{\mathcal{X}})$, we have
\begin{equation}
    \begin{split}
         \log & p_{\theta}(y=0|\mathbf{x}) - \log p_{\theta^{old}}(y=0|\mathbf{x}) = \\
         & \ell_{ELBO}(q,\theta) - \ell_{ELBO}(q,\theta^{old}) + \\
         & KL[q(\omega)||p_{\theta}(\omega|\mathcal{P}_{\mathcal{X}})] - KL[q(\omega)||p_{\theta^{old}}(\omega|\mathcal{P}_{\mathcal{X}})].
    \end{split}
\end{equation}
Since $KL[q(\omega)||p_{\theta}(\omega|\mathcal{P}_{\mathcal{X}})] \ge KL[q(\omega)||p_{\theta^{old}}(\omega|\mathcal{P}_{\mathcal{X}})]$ and that
$\ell_{ELBO}(q,\theta) - \ell_{ELBO}(q,\theta^{old}) =$ \\
\scalebox{0.88}{
        $\mathbb{E}_{q(\omega)}[\log p_{\theta}(y=0,\omega|\mathbf{x},\mathcal{P}_{\mathcal{X}})] - 
    \mathbb{E}_{q(\omega)}[\log p_{\theta^{old}}(y=0,\omega|\mathbf{x},\mathcal{P}_{\mathcal{X}})]$,} 
we conclude that 
\begin{equation}
\begin{split}
\log  & p_{\theta}(y=0|\mathbf{x},\mathcal{P}_{\mathcal{X}}) -  \log p_{\theta^{old}}(y=0|\mathbf{x},\mathcal{P}_{\mathcal{X}})  \ge \\  
&\mathbb{E}_{q(\omega)}[\log p_{\theta}(y=0,\omega|\mathbf{x},\mathcal{P}_{\mathcal{X}})] - \\ 
    &\mathbb{E}_{q(\omega)}[\log p_{\theta^{old}}(y=0,\omega|\mathbf{x},\mathcal{P}_{\mathcal{X}})] \ge 0        
\end{split}
\label{eq:difference_bound}
\end{equation}
because of the assumption in this Lemma.
\end{proof}

\section{Convergence Conditions Proof}
\label{sec:convergence_analysis_proof}

\begin{restatable}[]{lemma}{Convergence}
    \label{thm:convergence}
    Assume that $\{ \theta^{(e)}\}_{e=1}^{+\infty}$ denotes the sequence of trained model parameters from the constrained optimisation of $\ell_{ELBO}$ such that: 1) the sequence $\{ \log p_{\theta^{(e)}}(y=0|\mathbf{x},\mathcal{P}_{\mathcal{X}})\}_{e=1}^{+\infty}$ is bounded above, and 2) 
    \scalebox{0.79}{$\left(\mathbb{E}_{q(\omega)}[\log p_{\theta^{(e+1)}}(y=0,\omega|\mathbf{x},\mathcal{P}_{\mathcal{X}})] - 
    \mathbb{E}_{q(\omega)}[\log p_{\theta^{(e)}}(y=0,\omega|\mathbf{x},\mathcal{P}_{\mathcal{X}})]\right) \ge $} \\
    \scalebox{0.8}{
    $\xi \left(\theta^{(e+1)}-\theta^{(e)}\right)^{\top}\left(\theta^{(e+1)}-\theta^{(e)}\right)$}, 
    for $\xi>0$ and all $e \ge 1$, and $q(\omega)=p_{\theta^{(e)}}(\omega|\mathcal{P}_{\mathcal{X}})$.  
    Then $\{\theta^{(e)}\}_{e=1}^{+\infty}$ converges to some $\theta^{\star} \in \Theta$.
 \end{restatable}
\begin{proof}
We follow the proof for Theorem 2 in~\citep{dempster1977maximum}.
The sequence $\{ \log p_{\theta^{(e)}}(y=0|\mathbf{x},\mathcal{P}_{\mathcal{X}})\}_{e=1}^{+\infty}$ is non-decreasing (from Lemma~\ref{thm:correctness}) and bounded  above (from assumption (1) in Lemma~\ref{thm:convergence}), so it converges to $L^{\star} < +\infty$. 
Hence, using Cauchy criterion~\citep{nguyen2020tutorial}, for any $\epsilon > 0$, we have $e^{(\epsilon)}$ such that, for $e \ge e^{(\epsilon)}$ and all $r \ge 1$,
\begin{equation}
\begin{split}
    \sum_{j=1}^r & \left( \log p_{\theta^{(e+j)}}(y=0|\mathbf{x},\mathcal{P}_{\mathcal{X}}) - \log p_{\theta^{(e+j-1)}}(y=0|\mathbf{x},\mathcal{P}_{\mathcal{X}}) \right) = \\
    & \left( \log p_{\theta^{(e+r)}}(y=0|\mathbf{x},\mathcal{P}_{\mathcal{X}}) - \log p_{\theta^{(e)}}(y=0|\mathbf{x},\mathcal{P}_{\mathcal{X}}) \right) < \epsilon.
\end{split}
    \label{eq:bound_1_thm2}
\end{equation}
From~\eqref{eq:difference_bound},
\begin{equation}
\begin{split}
    0 & \le 
    \mathbb{E}_{q(\omega)}[\log p_{\theta^{(e+j)}}(y=0,\omega|\mathbf{x},\mathcal{P}_{\mathcal{X}})] - \\ 
    & \;\;\;\;\;\; \mathbb{E}_{q(\omega)}[\log p_{\theta^{(e+j-1)}}(y=0,\omega|\mathbf{x},\mathcal{P}_{\mathcal{X}})] \\
    & \le
    \log p_{\theta^{(e+j)}}(y=0|\mathbf{x},\mathcal{P}_{\mathcal{X}}) - \log p_{\theta^{(e+j-1)}}(y=0|\mathbf{x},\mathcal{P}_{\mathcal{X}})
\end{split}
    \label{eq:bound_2_thm2}
\end{equation}
for $j \ge 1$ and $q(\omega)=p_{\theta^{(e+j-1)}}(\omega|\mathcal{P}_{\mathcal{X}})$. Hence, from \eqref{eq:bound_1_thm2},
\begin{equation}
\begin{split}
    \sum_{j=1}^r & ( \mathbb{E}_{q(\omega)}[\log p_{\theta^{(e+j)}}(y=0,\omega|\mathbf{x},\mathcal{P}_{\mathcal{X}}))] - \\
    & \mathbb{E}_{q(\omega)}[\log p_{\theta^{(e+j-1)}}(y=0,z|\mathbf{x},\mathcal{P}_{\mathcal{X}}))] ) < \epsilon,
\end{split}
    \label{eq:bound_3_thm2}
\end{equation}
for $e \ge e^{(\epsilon)}$ and all $r \ge 1$.
Given assumption (2) in Lemma~\ref{thm:convergence} for $e,e+1,e+2,...,e+r-1$, we
 have from \eqref{eq:bound_3_thm2},
 \begin{equation}
     \epsilon > \xi \sum_{j=1}^{r} \left( \theta^{(e+j)} - \theta^{(e+j-1)} \right)^{\top}\left( \theta^{(e+j)} - \theta^{(e+j-1)} \right),
 \end{equation}
so
\begin{equation}
     \epsilon > \xi \left( \theta^{(e+r)} - \theta^{(e)} \right)^{\top}\left( \theta^{(e+r)} - \theta^{(e)} \right),
 \end{equation}
 which is a requirement to prove the convergence of $\theta^{(e)}$ to some $\theta^{\star} \in \Theta$.
\end{proof}

\bgroup
\def\arraystretch{1.4}
\begin{table*}[t!]
\centering
\resizebox{0.95\linewidth}{!}{%
\begin{tabular}{@{}ccccccccccccc@{}}
\toprule
        &Method                                        &	0   &	1	    &	2	    &	3	    &	4	    &	5	    &	6	    &	7	    &	8	    &	9	    &	Mean	\\	\midrule\midrule
        &DAE~\cite{hadsell2006dimensionality} &	0.894	        &	0.999	&	0.792	&	0.851	&	0.888	&	0.819	&	0.944	&	0.922	&	0.740	&	0.917	&	0.8766	\\	\cmidrule{2-13}
        &VAE~\cite{vae}  &	0.997	        &	0.999	&	0.936	&	0.959	&	0.973	&	0.964	&	0.993	&	0.976	&	0.923	&	0.976	&	0.9696	\\	\cmidrule{2-13}
        &KDE~\cite{bishop2006pattern}&	0.885	        &	0.996	&	0.710	&	0.693	&	0.844	&	0.776	&	0.861	&	0.884	&	0.669	&	0.825	&	0.8140	\\	\cmidrule{2-13}
        &OCSVM~\cite{oc-svm}	                        &	0.988	        &	0.999	&	0.902	&	0.950	&	0.955	&	0.968	&	0.978	&	0.965	&	0.853	&	0.955	&	0.9510	\\	\cmidrule{2-13}
        &AnoGAN~\cite{anogan}	                        &	0.966	        &	0.992	&	0.850	&	0.887	&	0.894	&	0.883	&	0.947	&	0.935	&	0.849	&	0.924	&	0.9127	\\	\cmidrule{2-13}
        &DSVDD~\cite{dsvdd}                        &	0.980	        &	0.997	&	0.917	&	0.919	&	0.949	&	0.885	&	0.983	&	0.946	&	0.939	&	0.965	&	0.9480	\\	\cmidrule{2-13}
      &OCGAN~\cite{ocgan}	                        &	0.998	        &	0.999	&	0.942	&	0.963	&	0.975	&	0.980	&	0.991	&	0.981	&	0.939	&	0.981	&	0.9750	\\	\cmidrule{2-13}
        &PixelCNN~\cite{van2016conditional}	                    &	0.531	        &	0.995	&	0.476	&	0.517	&	0.739	&	0.542	&	0.592	&	0.789	&	0.340	&	0.662	&	0.6180	\\	\cmidrule{2-13}
        &CapsNet\textsubscript{PP}~\cite{li2020exploring}	    &	0.998	        &	0.990	&	0.984	&	0.976	&	0.935	&	0.970	&	0.942	&	0.987	&	\textbf{0.993}	&	0.990	&	0.9770	\\	\cmidrule{2-13}
        &CapsNet\textsubscript{RE}~\cite{li2020exploring}	    &	0.947	        &	0.907	&	0.970	&	0.949	&	0.872	&	0.966	&	0.909	&	0.934	&	0.929	&	0.871	&	0.9250	\\	\cmidrule{2-13}
        &ADGAN~\cite{ADGAN}	                        &	\textbf{0.999}	&	0.992	&	0.968	&	0.953	&	0.960	&	0.955	&	0.980	&	0.950	&	0.959	&	0.965	&	0.9680	\\	\cmidrule{2-13} %\cmidrule{2-13}
        &LSA~\cite{lsa}	                            &	0.993	        &	0.999	&	0.959	&	0.966	&	0.956	&	0.964	&	0.994	&	0.980	&	0.953	&	0.981	&	0.9750	\\	\cmidrule{2-13}
        % MemAE	&		&		&		&		&		&		&		&		&		&		&		\\	\hline
        &GradCon~\cite{gradcon}	                        &	0.995	        &	0.999	&	0.952	&	0.973	&	0.969	&	0.977	&	0.994	&	0.979	&	0.919	&	0.973	&	0.9730	\\	\cmidrule{2-13}
        &$\lambda$-VAE\textsubscript{u}~\cite{lamda-vae}	&	0.991	        &	0.996	&	0.983	&	0.978	&	0.976	&	0.972	&	0.993	&	0.981	&	0.98	&	0.967	&	0.9820	\\	\cmidrule{2-13}
        &ULSLM~\cite{ulslm}	                        &	0.991	        &	0.972	&	0.919	&	0.943	&	0.942	&	0.872	&	0.988	&	0.939	&	0.96	&	0.967	&	0.9490	\\	\cmidrule{2-13}
        &CAVGA-D\textsubscript{u}~\cite{venkataramanan2019attention}	    &	0.994	        &	0.997	&	0.989	&	0.983	&	0.977	&	0.968	&	0.988	&	0.986	&	0.988	&	\textbf{0.991}	&	0.9860	\\	\cmidrule{2-13}
    	&	Student-Teacher~\cite{bergmann2020uninformed}	&	\textbf{0.999}	&	0.999	&	0.990	&	\textbf{0.993}	&	0.992	&	\textbf{0.993}	&	0.997	&	\textbf{0.995}	&	\textbf{0.986}	&	0.991	&	\textbf{0.9935}	\\	\cmidrule{2-13}
    	                        &	\textbf{Ours - ImageNet}                                       	&	0.998	&	\textbf{0.999}	&	\textbf{0.992}	&	0.991	&	\textbf{0.993}	&	0.991	&	\textbf{0.997}	&	0.990	&	0.984	&	\textbf{0.991}	&	\textbf{0.9927}	\\	\bottomrule

\end{tabular}}
\caption{\textbf{Anomaly detection:} class-level testing AUC on MNIST produced by the SOTA and our methods.}
\label{tab:mnist-auc}
\end{table*}

\bgroup
\def\arraystretch{1.4}
\begin{table*}[h]
\centering
\resizebox{0.95\linewidth}{!}{%
\begin{tabular}{@{}c|cccccccccccc@{}}
\toprule
	Method	&	0	&	1	&	2	&	3	&	4	&	5	&	6	&	7	&	8	&	9	&	Mean	\\	\midrule
	\midrule
	Ours - ImageNet	&	0.908	&	0.992	&	0.902	&	0.946	&	0.93	&	0.95	&	0.818	&	0.993	&	0.938	&	0.981	&	0.935	\\
	Ours - SSL	&	0.926	&	0.992	&	0.922	&	0.946	&	0.931	&	0.971	&	0.832	&	0.992	&	0.946	&	0.982	&	0.944	\\\bottomrule
\end{tabular}}
\caption{\textbf{Anomaly detection:} class-level testing AUC on FMNIST produced by our methods.}
\label{tab:fmnist-auc}
\end{table*}

\section{Class-level Results}
The class-level results are shown in Tables~\ref{tab:auc_class_mvtec},~\ref{tab:mnist-auc},~\ref{tab:fmnist-auc},~\ref{tab:cifar-auc},  and~\ref{tab:mvtec-loc-auc}.  
%\yuanhong{Do we need to discuss all the numbers in details here or we just refer to the table?}
The mean accuracy and class-level anomaly detection accuracy on MVTec dataset is displayed in Tab.~\ref{tab:auc_class_mvtec}, where our ImageNet KD pre-trained model outperforms the previous SOTA methods CAVGA-D\textsubscript{u} and CAVGA-R\textsubscript{u}~\cite{venkataramanan2019attention} by 6\% and 2\%, respectively, and our SSL pre-trained model outperforms their approach by 7\% and 3\%, respectively. 
With ImageNet KD pre-training, our model achieves the best accuracy results in \textbf{ten categories} of the MVTec AD. The shallow generative baselines, such as DAE, AE-SSIM and AnoGAN yield sub-optimal results on MVTec AD. When compared with methods recently considered to be the MVTec AD SOTA, such as LSA~\cite{lsa} and $\lambda$-VAE\textsubscript{u}~\cite{lamda-vae}, our approach shows more than 7\% improvement. 
We also show the AUC anomaly detection results in Tab.~\ref{tab:auc_class_mvtec}, where our method, with SSL and ImageNet KD pre-training, surpasses all previous methods by at least 5.3\%, and produces the best results in eleven categories.  
%In Tables~\ref{tab:auc_class_mvtec}, we note that self-supervised learning approaches generally do not work well without a powerful anomaly detector for real-world anomaly images, such as the ones on MVTec AD~\cite{bergman2020classification,golan2018deep}. However, combining our IGD with the SSL pre-training, our model achieves the highest 93.4\% mean AUC and 85\% mean accuracy results and surpass the previous SOTA methods.
The results of IGD for MNIST in Tab.\ref{tab:mnist-auc} show that our approach pre-trained with ImageNet KD is competitive with the Student-Teacher~\cite{bergmann2020uninformed}, and both are better than any of the previously proposed methods in the field. 
In Table~\ref{tab:fmnist-auc}, we only show the results of our approach because we could not find the class-level results for other approaches.
On the class-level results for CIFAR10, on Tab.~\ref{tab:cifar-auc}, we notice that our approach pre-trained with ImageNet and SSL shows the best AUC result in the field by a large margin (around 10\%) compared with the Student-Teacher~\cite{bergmann2020uninformed} approach. Finally, the class-level anomaly localisation AUC results for MVTec on Tab.~\ref{tab:mvtec-loc-auc} only shows the results of our approach because we could not find results from other approaches.

\begin{table*}[t]
\centering
\resizebox{0.95\linewidth}{!}{%
\begin{tabular}{@{}ccccccccccccc@{}}
\toprule
	                            &	Method	&	Plane	&	Car	    &	Bird	&	Cat	&	Deer	&	Dog	&	Frog	&	Horse	&	Ship	&	Truck	&	Mean	\\	\midrule
	&	DAE~\cite{hadsell2006dimensionality}	                            &	0.411	&	0.478	&	0.616	&	0.562	&	0.728	&	0.513	&	0.688	&	0.497	&	0.487	&	0.378	&	0.5358	\\	\cmidrule{2-13}
	&	VAE~\cite{vae}	                            &	0.634	&	0.442	&	0.640	&	0.497	&	0.743	&	0.515	&	0.745	&	0.527	&	0.674	&	0.416	&	0.5833	\\	\cmidrule{2-13}
	&	 KDE~\cite{bishop2006pattern}	                        &	0.658	&	0.520	&	0.657	&	0.497	&	0.727	&	0.496	&	0.758	&	0.564	&	0.680	&	0.540	&	0.6100	\\	\cmidrule{2-13}
	&	OCSVM~\cite{oc-svm}	                        &	0.630	&	0.440	&	0.649	&	0.487	&	0.735	&	0.500	&	0.725	&	0.533	&	0.649	&	0.508	&	0.5860	\\	\cmidrule{2-13}
	&	AnoGAN~\cite{anogan}	                        &	0.671	&	0.547	&	0.529	&	0.545	&	0.651	&	0.603	&	0.585	&	0.625	&	0.758	&	0.665	&	0.6179	\\	\cmidrule{2-13}
	&	DSVDD~\cite{dsvdd}	                        &	0.617	&	0.659	&	0.508	&	0.591	&	0.609	&	0.657	&	0.677	&	0.673	&	0.759	&	0.731	&	0.6481	\\	\cmidrule{2-13}
    &	OCGAN~\cite{ocgan}	&	0.757	&	0.531	&	0.640	&	0.62	&	0.723	&	0.620	&	0.723	&	0.575	&	0.820	&	0.554	&	0.6566	\\	\cmidrule{2-13}
	&	PixelCNN~\cite{van2016conditional}	                    &	0.788	&	0.428	&	0.617	&	0.574	&	0.511	&	0.571	&	0.422	&	0.454	&	0.715	&	0.426	&	0.5510	\\	\cmidrule{2-13}
	&	CapsNet\textsubscript{PP}~\cite{li2020exploring}	    &	0.622	&	0.455	&	0.671	&	0.675	&	0.683	&	0.350	&	0.727	&	0.673	&	0.710	&	0.466	&	0.6120	\\	\cmidrule{2-13}
	&	CapsNet\textsubscript{RE}~\cite{li2020exploring}	    &	0.371	&	0.737	&	0.421	&	0.588	&	0.388	&	0.601	&	0.491	&	0.631	&	0.410	&	0.671	&	0.5310	\\	\cmidrule{2-13}
	&	ADGAN~\cite{ADGAN}	                        &	0.671	&	0.547	&	0.529	&	0.545	&	0.651	&	0.603	&	0.585	&	0.625	&	0.758	&	0.665	&	0.6180	\\	\cmidrule{2-13}
	&	LSA~\cite{lsa}	                            &	0.735	&	0.580	&	0.690	&	0.542	&	0.761	&	0.546	&	0.751	&	0.535	&	0.717	&	0.548	&	0.6410	\\	\cmidrule{2-13}
% 	&	MemAE	&		&		&		&		&		&		&		&		&		&		&		\\	\cmidrule{2-13}
	&	GradCon~\cite{gradcon}	                        &	0.760	&	0.598	&	0.648	&	0.586	&	0.733	&	0.603	&	0.684	&	0.567	&	0.784	&	0.678	&	0.6640	\\	\cmidrule{2-13}
	&	$\lambda$-VAE\textsubscript{u}~\cite{lamda-vae}	&	0.702	&	0.663	&	0.68	&	0.713	&	0.77	&	0.689	&	0.805	&	0.588	&	0.813	&	0.744	&	0.7170	\\	\cmidrule{2-13}
	&	ULSLM~\cite{ulslm}	                        &	0.740	&	0.747	&	0.628	&	0.572	&	0.678	&	0.602	&	0.753	&	0.685	&	0.781	&	0.795	&	0.7360	\\	\cmidrule{2-13}
	&	CAVGA-D\textsubscript{u}~\cite{venkataramanan2019attention}	    &	0.653       	&	0.784	        &	\textbf{0.761}	    &	\textbf{0.747}	&	0.775	&	0.552	&	0.813	&	0.745	&	0.701	&	0.741	&	0.7370	\\	\cmidrule{2-13}
	&	Student-Teacher~\cite{bergmann2020uninformed}	&	0.789	&	0.849	&	0.734	&	\textbf{0.748}	&	\textbf{0.851}	&	\textbf{0.793}	&	\textbf{0.892}	&	0.830	&	0.862	&	0.848	&	0.8196	\\	\cmidrule{2-13}
	                        &\textbf{Ours - ImageNet}	        &	\textbf{0.868}	&	\textbf{0.870}	&	\textbf{0.738}	&	0.716	&	0.850	&	0.766	&	0.890	&	\textbf{0.871}	&	\textbf{0.898}	&	\textbf{0.899}	&	\textbf{0.8368}
	                        \\ \cmidrule{2-13}
	                         &\textbf{Ours - SSL}	         &	\textbf{0.906}	&	\textbf{0.979}	&	\textbf{0.839}	&	\textbf{0.823}&	\textbf{0.886}	&	\textbf{0.899}	&	\textbf{0.909}	&	\textbf{0.964}	&	\textbf{0.969}	&	\textbf{0.948}	&	\textbf{0.9125} \\ \bottomrule

\end{tabular}}
\caption{\textbf{Anomaly detection:} class-level testing AUC on CIFAR10 produced by the SOTA and our methods.}
\label{tab:cifar-auc}
\end{table*}

\begin{table*}[t]
\centering
\resizebox{0.95\linewidth}{!}{%
\begin{tabular}{@{}c|ccccccccccccccccc@{}}
\toprule
Method     & Bottle & Hazelnut & Capsule & Metal Nut & Leather & Pill & Wood & Carpet & Tile & Grid & Cable & Transistor & Toothbrush & Screw & Zipper & Mean \\\midrule
Ours - ImageNet  & 0.928  & 0.981 & 0.967 & 0.902 & 0.983 & 0.962 & 0.827 & 0.901 & 0.727 & 0.916 & 0.835 & 0.843 & 0.974 & 0.960 & 0.932 & 0.909 \\
Ours - SSL & 0.922  & 0.980  & 0.977 & 0.926 & 0.995 & 0.973 & 0.891 & 0.947 & 0.780  & 0.977 & 0.847 & 0.844 & 0.977 & 0.970 & 0.967 & 0.931 \\\bottomrule
\end{tabular}}
\caption{\textbf{Anomaly localisation:} class-level testing pixel-wise localisation AUC results on the anomalous images of MVTec AD produced by our methods.}
\label{tab:mvtec-loc-auc}
\end{table*}
%\clearpage

\section{Qualitative Localisation Results}

Figure~\ref{fig:colon_visual} shows the polyp  segmentation results on Hyper-Kvasir testing set images, and Figure~\ref{fig:my_label} displays the defect results on MVTec AD testing set images.

\begin{figure*}[h!]
  \centering
    \includegraphics[width=.95\textwidth]{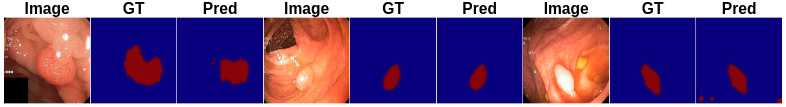}
    \caption{\label{fig:colon_visual}
    Qualitative visual results from Hyper-Kvasir testing set (red = anomaly). 
    } 
\end{figure*}

\begin{figure*}[h!]
    \centering
    \includegraphics[width=0.95\linewidth]{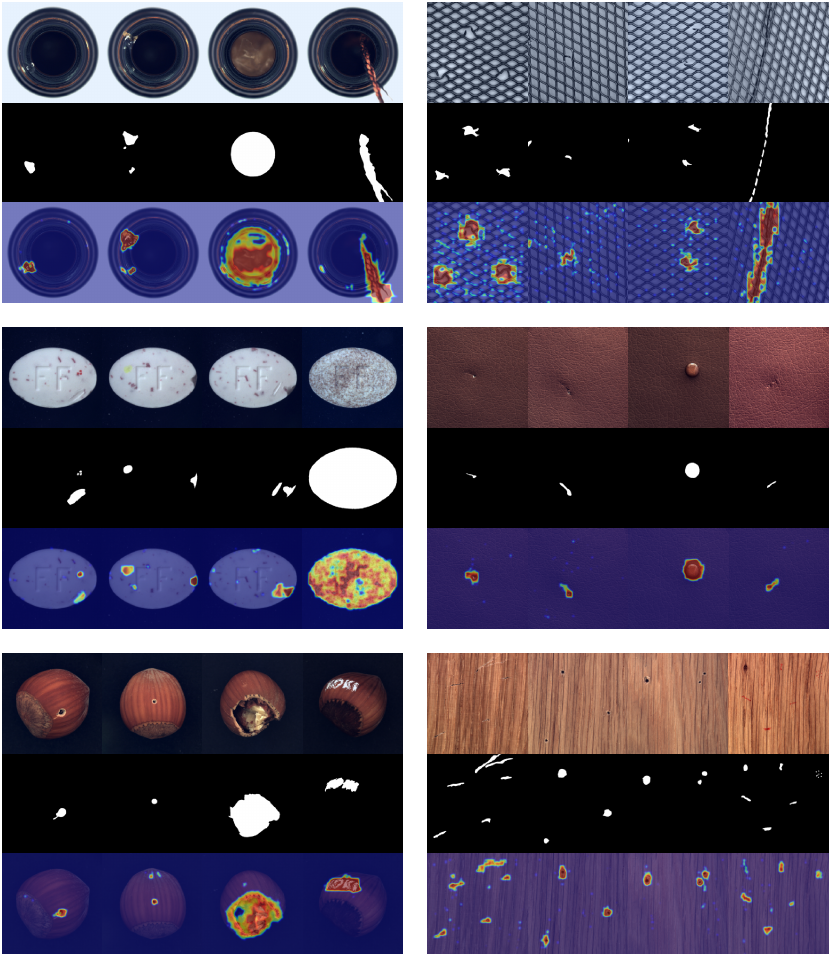}
    \caption{Qualitative results of our anomaly localisation results on the MVTec AD testing set (red = high probability of anomaly).}
    \label{fig:my_label}
\end{figure*}

\end{document}